 \documentclass[preprint,3p,authoryear]{elsarticle}
\usepackage{natbib} 

\usepackage{booktabs}
\usepackage{url}
\usepackage{graphicx}
\usepackage{balance}
\usepackage{amsmath}
\usepackage{placeins}
\usepackage{subfigure}
\usepackage{graphicx}
\usepackage{balance}
\usepackage{color}
\usepackage{comment}
\usepackage{array}
\usepackage{lipsum}
\usepackage{caption}
\usepackage[bookmarks,bookmarksnumbered]{hyperref} 
\hypersetup{colorlinks = true,
    linkcolor=blue,
    filecolor=magenta,      
    urlcolor=cyan}

\usepackage{blindtext}
\usepackage{enumitem}
\usepackage{lipsum}
\usepackage{capt-of}
\usepackage{tikz}
\usetikzlibrary{arrows, positioning, automata}
\usepackage{balance}
\usepackage{amsmath}
\usepackage[T1]{fontenc}
\usepackage{algorithm}
\usepackage{algpseudocode}
\usepackage{subfigure}
\usepackage{graphicx}
\usepackage[utf8]{inputenc}
\usepackage{balance}
\usepackage{multirow}
\usepackage{multicol}
\usepackage{setspace}
\usepackage{color}
\usepackage{comment}
\usepackage{array}
\usepackage{soul}
\usepackage[justification=centering]{caption}
\usepackage[lighttt]{lmodern}
\usepackage{libertine}
\usepackage{amssymb}
\usepackage{float}
\usepackage{hyperref}

\usepackage{flushend} 
\usepackage[table]{xcolor}
\usepackage{adjustbox}

\begin{document}

\begin{titlepage}
\begin{center}
\vspace*{10pt}
\doublespacing
{\Large Uncertainty-Aware Semi-Supervised Framework with Supervised Contrastive Learning for Sugarcane Leaf Disease Detection}
\vspace*{25pt}

Shalini Dangi$^{a}$ (phd2301201003@iiti.ac.in), Divya Teja$^{a}$ (cse210001040@iiti.ac.in), \\ Omkar Kabde$^{a}$ (ugs22058\_cic.omkar@cbit.org.in), Shahid Shafi Dar$^{a}$ (phd2201201004@iiti.ac.in), \\Nagendra Kumar$^a$ (nagendra@iiti.ac.in) \\

\hspace{5pt}




\section*{Highlights}

\begin{itemize}
    \item An uncertainty-aware semi-supervised network is proposed for sugarcane leaf disease detection.
    
    \item Discrete wavelet transform is used to enhance disease-related image features.
    
    \item KL-divergence-based refinement improves the reliability of pseudo-labels.
    
    \item Contrastive learning strengthens class-discriminative feature representations.
    
    \item Experimental results show effective learning from limited annotations and competitive performance.
\end{itemize}

\vspace{1em}

\begin{flushleft}
This is the preprint version of the accepted paper.

This paper is accepted in \textbf{ISPRS Journal of Photogrammetry and Remote Sensing}, 2026.

DOI: \url{https://doi.org/10.1016/j.isprsjprs.2026.06.005}
\end{flushleft}

\end{center}
\end{titlepage}

\begin{frontmatter}
\title{Uncertainty-Aware Semi-Supervised Framework with Supervised Contrastive Learning for Sugarcane Leaf Disease Detection}

\author[1]{Shalini Dangi}
\ead{phd2301201003@iiti.ac.in}

\author[1]{Divya Teja}
\ead{cse210001040@iiti.ac.in}

\author[2]{Omkar Kabde}
\ead{ugs22058_cic.omkar@cbit.org.in}

\author[1]{Shahid Shafi Dar}
\ead{phd2201201004@iiti.ac.in}

\author[1]{Nagendra Kumar \corref{cor1}}
\cortext[cor1]{Corresponding author email: \textit{nagendra@iiti.ac.in}}

\address[1]{Indian Institute of Technology Indore, Madhya Pradesh, India}

\address[2]{Chaitanya Bharathi Institute of Technology, Telangana, India}

\begin{abstract}
Sugarcane leaf diseases severely impact crop yield and quality, making accurate and timely detection critical for sustainable agriculture. Conventional deep learning models for plant disease detection depend on large annotated datasets, which are expensive to obtain and challenging to prepare due to challenges commonly observed across crop disease detection tasks, including (i) high intra-class variation caused by diverse symptom expressions across growth stages, (ii) subtle inter-class similarities between diseased and healthy tissues, and (iii) the risk of noisy labels degrading model reliability. To handle these challenges, we presented an \textbf{U}ncertainty-aware Mean-Teacher based \textbf{S}emi-\textbf{S}upervised Learning \textbf{D}isease Detection \textbf{Net}work (U\text{-}SSDNet) that leverages unlabeled data while maintaining robustness. First, discrete wavelet transform preprocessing is applied to enhance lesion visibility and reduce intra-class variation. Subsequently, the KL divergence is employed to refine the pseudo-labels and reduce the impact of confirmation bias. At the same time, supervised contrastive learning is applied to make the features more distinguishable and improve the separation among different classes.  
Experimental evaluations on sugarcane leaf disease datasets, considered a representative case study, demonstrate that the proposed approach achieves performance close to that of fully supervised models while requiring far fewer annotated samples and consistently outperforms state-of-the-art supervised and semi-supervised approaches.
\end{abstract}

\begin{keyword}
 Precision Agriculture, Plant disease detection, Semi-supervised learning, Consistency regularization, Contrastive learning 
\end{keyword}
\end{frontmatter}

\section{Introduction}
\label{sec:sa_int}
Agriculture sustains global food supply, rural life, and economic strength. With continued global population growth, the need for food and renewable resources increases, putting significant pressure on farming to produce more without harming the environment. Sugarcane is also an important crop worldwide, accounting for about 80\% of global sugar production. However, cultivating sugarcane is challenging because many diseases attack the plants, leading to significant reductions in crop quality and yield. Leaf diseases can affect plant health by reducing growth rates and lowering sugar content, posing major challenges for both farmers and the sugar production sector \citep{Barbedo_2016}. These leaf diseases spread rapidly, making early detection difficult in favorable weather. 
Although laboratory methods like PCR and DNA sequencing are highly accurate and require substantial resources, making them less practical for large-scale use \citep{Strachan_2022}.

Conventional disease classification methods depend on biochemical tests and expert-based visual inspection. Although accurate, biochemical methods are costly, require chemical reagents and specialized labor, and cannot be easily scaled to large fields. On the other hand, Manual inspection is labor-intensive and impractical for monitoring large-scale farms. Computer vision and deep learning approaches have been used to overcome these constraints in leaf disease detection. Different Convolutional Neural Network (CNN) architectures such as Inception-V3, ResNet50, VGG16, and VGG19 have shown notable performance in extracting features from plant leaf images ~\citep{Dey_2022}. 
The performance of CNN-based models mostly depends on the availability of labeled datasets. In practice, plant disease detection tasks commonly suffer from high intra-class variation, where the same disease may appear differently across growth stages and environmental conditions, as well as subtle inter-class similarities, where different diseases exhibit overlapping visual patterns with healthy tissues~\citep{Lu_2020, Barbedo_2018}. These challenges are widely observed across crop disease detection tasks and are not specific to a single crop. Variations in symptom appearance and real-field imaging conditions in sugarcane leaf disease detection intensify these challenges. As a result, limited feature diversity and a higher risk of overfitting can restrict the practical use of deep learning models.

Limited labeled agricultural data is a major challenge for building accurate plant disease detection models, as expert annotation is costly and time-consuming. Semi-supervised learning (SSL) offers a promising solution by leveraging large amounts of unlabeled data alongside limited labeled samples to improve generalization and robustness. For example, SSL has enhanced performance with ConvNeXt models using consistency regularization for precise weed detection~\citep{Benchallal_2024}, and attention-based self-supervised contrastive learning has improved plant disease detection by extracting diverse leaf disease features from unlabeled images~\citep{Yilma_2025, Dangi_2025}. Frameworks that incorporate both intra-class and inter-class information have also improved pest identification in real-field scenarios~\citep{Hu_2025}, and uncertainty-aware optimization in semi-supervised contrastive learning has increased robustness in agricultural monitoring~\citep{Ai_2025}. These findings demonstrate that SSL provides a scalable, cost-effective solution for plant disease detection with limited supervision. However, challenges related to robustness and uncertainty handling remain open research problems.

This study presents an SSL-based Mean-Teacher approach for detecting sugarcane leaf disease. The Mean-Teacher model improves performance by enforcing consistency between the student and teacher networks and leveraging unlabeled data. Sugarcane leaf images are processed using the Discrete Wavelet Transform (DWT) to improve feature representation and focus on patterns of leaf diseases. In addition, Kullback-Leibler divergence is used for pseudo-label correction to reduce confirmation bias and increase prediction accuracy. Supervised contrastive learning is also applied to improve class separation and disease detection performance. 
The proposed approach combines these techniques to address intra-class variation and inter-class similarity in plant disease detection, using sugarcane leaf disease as a representative case study.

The proposed research outlines the following main contributions:
\begin{itemize}
\item We propose a novel semi-supervised Mean-Teacher framework for plant disease detection, evaluated using sugarcane leaf disease datasets as a representative case study.
\item We introduce a discrete wavelet transform to enhance the semantic context of visual features, thereby improving the model's representation ability.
\item We leverage supervised contrastive learning across different leaf disease classes to improve class separability and enhance classification performance.
\item We propose a KL-divergence-based pseudo-label correction approach to reduce noise in pseudo-labels and improve model stability.
\end{itemize}

The rest of the manuscript is structured as follows: Section \ref{sec:sa_rw} details the literature works, Section \ref{Prob} define the problem formulation, Section \ref{Methodology} presents the detailed analysis of our proposed methodology, Section \ref{Results} describes the results analysis and the details of the experimental setup, Section \ref{Discussion} presents the discussions, including limitations, and Section \ref{Conclusion} presents the conclusion.

\section{Related Work} 
\label{sec:sa_rw}
\subsection{\textit{Traditional Methods}}
Plant leaf disease detection is a key challenge in modern agriculture. Early approaches mainly employ handcrafted feature extraction and machine learning classification methods such as Support Vector Machines (SVMs), Naive Bayes, Decision Trees, and Random Forests~\citep{Sari_2021, Mia_2020, Panigrahi_2020}. The above methods use handcrafted features such as color, shape, and texture to identify healthy and diseased leaves. These features can be affected by leaf position, background, changes in lighting, and inter-class variability. Their reliance on manual feature engineering introduces subjectivity and limits scalability, particularly in real-world conditions where data may be diverse and noisy. Therefore, their ability to generalize to unseen conditions or large-scale datasets remains limited. These challenges are not specific to a single crop and are commonly observed in plant disease detection under varied real-world conditions.

{\subsection{Deep Learning for Plant Disease Detection}}
Deep learning has emerged as a prevalent methodology for plant leaf disease detection due to the strong feature-learning capabilities of convolutional neural networks (CNNs). Early research demonstrated that CNN-based models are effective for identifying plant diseases, particularly under controlled imaging conditions~\citep{Dyrmann_2016, Mohanty_2016}. To further enhance classification performance, deeper network architectures such as ResNet and DenseNet were subsequently introduced, enabling richer feature representations and improved learning capacity ~\citep{Huang_2017, He_2016}. Comparative studies have consistently shown that CNN-based approaches outperform traditional machine learning techniques for crop disease recognition across multiple plant species ~\citep{Dey_2022, Shah_2022}. Despite their success, CNN-based models remain affected by high intra-class variation, subtle inter-class similarities, and environmental variability, which are common across plant disease detection tasks.
To address the challenge of limited annotated agricultural data, transfer learning has been widely adopted. In this approach, models pre-trained on large-scale datasets such as ImageNet are fine-tuned to improve plant leaf disease classification performance ~\citep{Lambor_2022, Singh_2022, Kavala_2022}. Furthermore, lightweight and computationally efficient architectures, including MobileNet and EfficientNet, have been explored to support deployment in resource-constrained agricultural environments \citep{Ali_2024, Liu_2022, Tang_2020}. However, prior studies indicate that the effectiveness of deep learning models is highly dependent on dataset size and diversity, and their performance often degrades under real-world field conditions ~\citep{ Barbedo_2018, Barbedo_2016}. Deep learning methods have also been applied to other agricultural tasks, such as query–response systems \citep{Rehman_2023} and yield prediction \citep{Dangi_2025}, to support farmers in decision-making and improve crop production.

{\subsection{Semi-Supervised Learning in Agriculture}}
{To mitigate the reliance on large labeled datasets, semi-supervised learning (SSL) has emerged as a promising paradigm in agricultural vision. SSL methods aim to exploit unlabeled data through techniques such as pseudo-labeling, self-training, and consistency regularization~\citep{Raghaw_2026}. Recent studies have shown that incorporating unlabeled samples can significantly improve plant disease recognition performance when labeled data are scarce~\citep{Benchallal_2024, Ilsever_2024}. These approaches are particularly suitable for agricultural applications, where collecting large-scale annotations is labor-intensive and costly~\citep{Thite_2024, Lu_2020}.
However, existing SSL methods face several challenges in practical agricultural scenarios. Pseudo-labeling approaches are prone to error accumulation due to incorrect predictions, while consistency-based methods often assume reliable predictions on unlabeled samples without explicitly modeling uncertainty~\citep{Hu_2025}. This limitation is critical for plant disease detection, where visual similarities between disease classes and variations in environmental conditions can lead to ambiguous predictions. Consequently, robust SSL frameworks that can handle noisy labels and uncertain samples remain an open research problem in agricultural vision.}

{\subsection{Mean-Teacher and Uncertainty-Aware Learning}}
The Mean-Teacher framework is a semi-supervised learning approach that employs a teacher-student architecture. The teacher network is updated as an exponential moving average (EMA) of the student network parameters, by enforcing consistency between student and teacher predictions on unlabeled data. Mean-Teacher models have shown enhanced generality and stability in different computer vision tasks~\citep{Tarvainen_2017}. This consistency-based learning strategy effectively regularizes the training process and reduces overfitting.
Despite its success, most Mean-Teacher-based approaches do not explicitly account for prediction uncertainty and treat all unlabeled samples equally during training. Recent research work in other application domains has shown that incorporating uncertainty-aware mechanisms, such as contrastive learning and uncertainty-weighted optimization, can significantly improve semi-supervised learning performance~\citep{Ai_2025, Liu_2024}. However, uncertainty-aware Mean-Teacher frameworks have rarely been explored in agricultural disease detection. Given the inherent variability, inter-class similarity, and noisy visual patterns in crop disease datasets~\citep{Ye_2024, Strachan_2022} and the lack of uncertainty modeling, this represents a critical gap that motivates the proposed approach. \\

\section{Problem Formulation}
\label{Prob}
We address semi-supervised image classification using both labeled and unlabeled data. 
The labeled dataset is defined in~\autoref{eq:labeled}:
\begin{equation}
\mathcal{D}_L = \{(x_i, y_i)\}_{i=1}^{N_L}
\label{eq:labeled}
\end{equation}
where \( x_i \) is an input image and \( y_i \in \{1, \ldots, C\} \) denotes its class label. 
The unlabeled dataset is given by ~\autoref{eq:Unlabeled}.
\begin{equation}
\mathcal{D}_U = \{x_j\}_{j=1}^{N_U}
\label{eq:Unlabeled}
\end{equation}
A teacher-student framework is employed, where the predictions of the teacher and student networks are represented by 
\( \hat{y}^T \) and \( \hat{y}^S \), respectively. 
The student is optimized using a supervised loss \( \mathcal{L}_{sup} \) on \( \mathcal{D}_L \) and an uncertainty-guided consistency loss 
\( \mathcal{L}_{cons} \) to align predictions on both \( \mathcal{D}_L \) and \( \mathcal{D}_U \). 
The total loss combines these two terms, enabling the model to utilize unlabeled data and improve generalization under limited annotations.

\section{Methodology}
\label{Methodology}
\begin{figure*}[h!]
    \centering
    \captionsetup{justification=justified, singlelinecheck=false}
    \includegraphics[width=\linewidth, height=9.5cm]{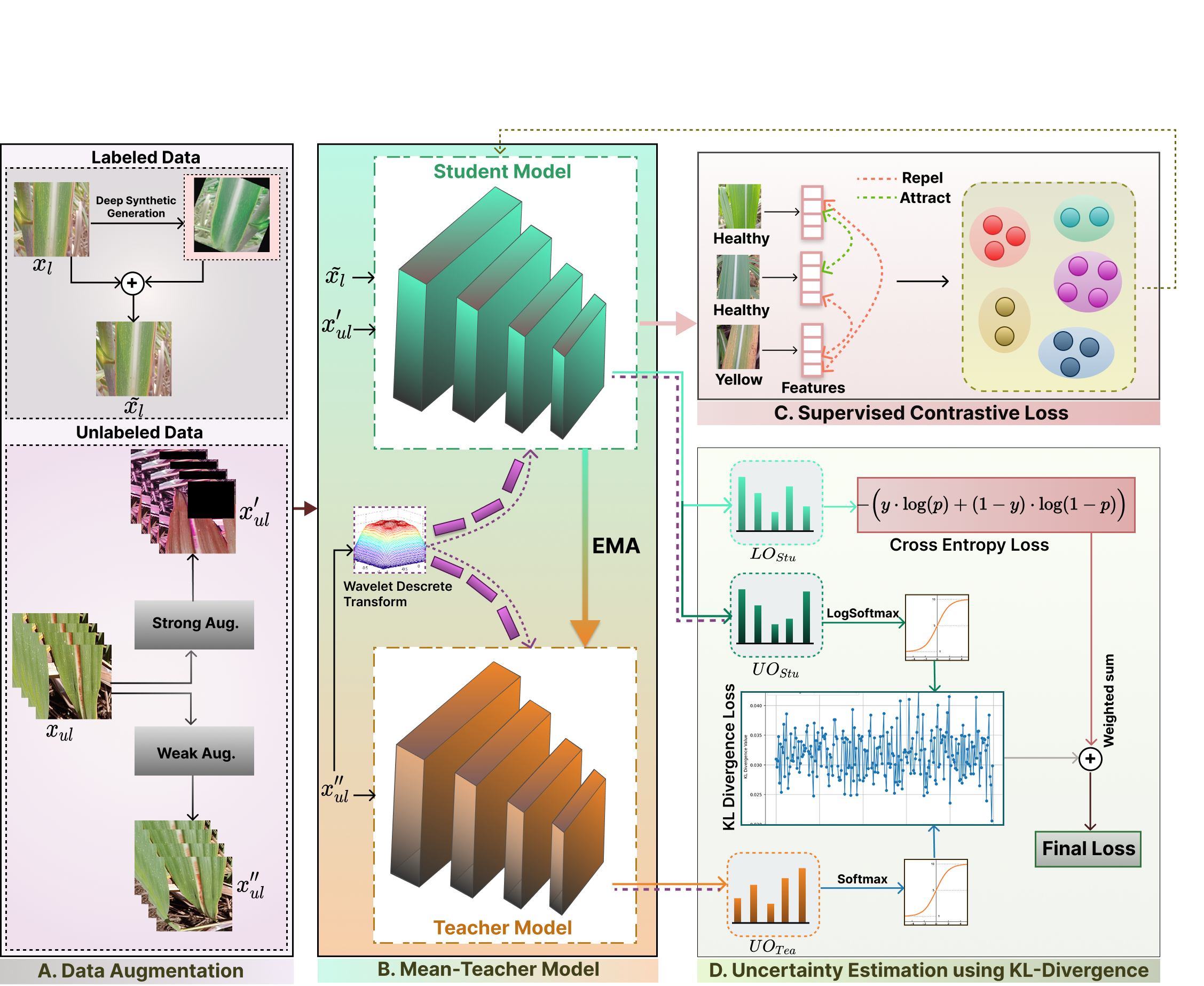}
    \caption{The proposed U-SSDNet framework contains four key elements: (i) Sugarcane Leaf Preprocessing and Augmentation; (ii) A Mean-Teacher model for Consistency Learning; (iii) Supervised Contrastive Learning for discriminative representation; and (iv) KL-divergence–based Uncertainty Estimation to guide the utilization of unlabeled data.}
    \label{fig:architecture}
\end{figure*}

The methodology section provides a complete overview of the proposed U-SSDNet framework for sugarcane leaf disease detection, as described in~\autoref{fig:architecture}. The framework includes four distinct sub-modules: (A) Sugarcane Leaf Preprocessing and Augmentation in which includes the initial processing such as resizing and augmentation; (B) Supervised Contrastive Learning for Inter-Class Discriminative Embedding, which optimizes feature representations by learning to contrast between different disease classes using a contrastive loss function in a supervised manner; (C) Mean-Teacher based Semi-Supervised Learning for Global Feature Representation, which utilizes a semi-supervised learning approach to enhance model generalization by uses both labeled and unlabeled data.

\subsection{\textit{Sugarcane Leaf Preprocessing and Augmentation:}} \label{section:social_context_feature}
The preprocessing stage includes resizing images and applying augmentation to increase data diversity and improve model performance. Data augmentation plays an essential role in leaf disease detection because the labeled dataset is often small and images vary in orientation and environmental conditions~\citep{Naveed_2024, Xu_2023}. It strengthens the model's robustness by replicating real-world aspects such as partial occlusion, changes in lighting,
and image blur. We first apply traditional augmentations, such as horizontal and vertical flips and rotations, to enable the model to generalize across different leaf positions and symmetries and introduce basic spatial variations. For labeled data, strong augmentations such as Color Jittering, Gaussian Blur, and Random Erasing are applied to create highly diverse training samples, which help improve the model's performance by providing it with a wide range 
of possible differences between diseased and healthy leaves.

\subsubsection{Color Jittering}
Color jittering introduces random perturbations to the image's color attributes, brightness, contrast, saturation, and hue to simulate various lighting and environmental conditions~\citep{Xu_2023}. Initially, the RGB image \( I \) is modified using a linear transformation to adjust brightness and contrast, as shown in~\autoref{eq:rgb_jitter}:

\begin{equation}
I' = T_{\text{jitter}}(I) = \alpha \cdot I + \beta
\label{eq:rgb_jitter}
\end{equation}

where \( \alpha \) is a random scalar controlling contrast, and \( \beta \) adjusts brightness.
To introduce additional variation, the image is converted to the HSV color space, allowing separate control over hue and saturation. The transformation in HSV space is given in ~\autoref{eq:hsv_jitter}, in which \( \Delta H \) represents the hue shift, and \( \alpha_s \) and \( \alpha_v \) are random scaling factors that modify saturation and brightness, respectively.

\begin{equation}
I'_{\text{HSV}} = \left[ H + \Delta H,\; S \cdot \alpha_s,\; V \cdot \alpha_v \right]
\label{eq:hsv_jitter}
\end{equation}

\subsubsection{Random Erasing}

Random Erasing is a data augmentation technique that enhances model robustness by simulating occlusions or missing regions in an image~\citep{Naveed_2024}. A random rectangular region \( R \subset I \) is selected, and the pixel values within this region are replaced as defined in Equation~\ref{eq:random_erasing}:

\begin{equation}
I'(x, y) =
\begin{cases}
v, & \text{if } (x, y) \in R, \\
I(x, y), & \text{otherwise},
\end{cases}
\label{eq:random_erasing}
\end{equation}

In our work with sugarcane leaf disease images, the region \( R \) is randomly positioned within the image, and its pixel values \( v \) are set to a fixed constant of 128, representing a neutral gray. The size of \( R \) is determined by an area ratio \( s \) between 0.02 and 0.33 relative to the image, and its width-to-height aspect ratio \( r \) is sampled from 0.3 to 3.3. Random Erasing is applied with a 50\% probability, ensuring meaningful occlusions that improve model robustness without obscuring critical lesion information.

\subsection{\textit{Supervised Contrastive Learning for Inter-Class Discriminative Embedding}} 
In sugarcane leaf disease detection, a key challenge is learning features that can differentiate between leaf diseases with similar visual appearances. Conventional feature extraction techniques often struggle to capture subtle differences among disease types, leading to overlapping features and reduced classification performance. This problem becomes more difficult under conditions such as complex backgrounds, uneven lighting, and limited labeled data, which reduce the model's ability to separate disease classes.

To address these issues, a Supervised Contrastive Learning (SCL) module is added to improve the discriminative ability of the EfficientNet-based feature extractor in the student model. By using labeled data, the SCL module makes features of the similar leaf disease class more compact while increasing the distance between features of different classes. 
In the proposed SCL module, contrastive pairs are constructed within each training mini-batch using only labeled samples. For a given anchor sample, all other samples in the batch that share the same class label are treated as positive pairs, while samples belonging to different classes are treated as negative pairs. The contrastive loss for each anchor is computed against all other labeled samples in the batch, encouraging compact feature representations for the same class and separation across different classes~\citep{Raghaw_2024}. Unlabeled samples are excluded from the SCL loss and are instead optimized through consistency regularization and pseudo-label correction within the Mean-Teacher framework.

{Following \cite{Chen_2020}, we implement SCL using a Siamese network. In this setup, two identical networks with shared weights process sample pairs to learn discriminative embeddings. Unlike many existing contrastive methods that are self-supervised or use pseudo-labeled data, our approach applies contrastive loss only to reliable labeled samples and integrates it within a Mean-Teacher framework. Labeled data improve class separation, while unlabeled data are handled via consistency regularization and uncertainty-aware pseudo-label refinement. This produces more stable and discriminative features under real-field conditions.
The objective of SCL is to bring feature vectors of the same class closer together while pushing apart vectors of different classes. For each anchor sample $i$, the loss is computed against all other labeled samples in the mini-batch of size $B_l$, encouraging compact representations for the same class and greater separation across classes, as shown in~\autoref{S_loss}.
\begin{equation}
\mathcal{L}_i = - \frac{1}{|\mathcal{S}(i)|} 
\sum_{j \in \mathcal{S}(i)} 
\log \left( 
\frac{\exp(\text{sim}(\mathbf{z}_i, \mathbf{z}_j) / \tau)}
{\sum_{k=1}^{B_l} \mathbf{1}_{[k \ne i]} \exp(\text{sim}(\mathbf{z}_i, \mathbf{z}_k) / \tau)} 
\right)
\label{S_loss}
\end{equation}}

{Here, $i$ is the anchor, $j \in \mathcal{S}(i)$ are the positive samples, and $k = 1, \dots, B_l$ indexes all labeled samples in the mini-batch. $\mathbf{z}_i$ and $\mathbf{z}_j$ are L2-normalized feature vectors from the student network, and $\text{sim}(\mathbf{z}_i, \mathbf{z}_j)$ denotes the cosine similarity. The temperature parameter $\tau$ controls the sharpness of feature separation: smaller values emphasize differences, while larger values smooth the similarity distribution.}
{The total SCL loss over the mini-batch is defined in~\autoref{SCL_loss}:
\begin{equation}
\mathcal{L}_{\text{SCL}} = \frac{1}{B_l} \sum_{i=1}^{B_l} \mathcal{L}_i
\label{SCL_loss}
\end{equation}}

Computing pairwise similarities for all labeled samples in a mini-batch of size $B_l$ results in a time complexity of $\mathcal{O}(B_l^2 d)$, where $d$ is the embedding dimension. The Mean-Teacher consistency regularization involves pointwise comparisons between student and teacher predictions for all samples in the batch, scaling linearly as $\mathcal{O}(B d)$, where $B$ is the total batch size (labeled + unlabeled). Overall, the per-iteration training complexity is dominated by the SCL component but remains efficient for practical mini-batch sizes.
By integrating SCL with the Mean-Teacher framework, the network learns discriminative, semantically meaningful features, improving classification accuracy and generalization for sugarcane leaf disease detection.

\subsection{\textit{Mean-Teacher based Semi-Supervised Learning for Global Feature Representation}} 
This Mean-Teacher-based feature extraction framework comprises four key components: feature enhancement, the Mean-Teacher model, pseudo-label rectification, and a unified overall loss formulation. The enhancement module leverages domain knowledge and data transformations to improve feature quality and preserve complex patterns before they enter the model. The core Mean-Teacher architecture facilitates semi-supervised learning by encouraging consistency between a student and its Exponential Moving Average (EMA) teacher under input, as shown in ~\autoref{fig:architecture}. A pseudo-label rectification mechanism is introduced to refine the learning cue from unlabeled data, correcting noisy or uncertain teacher predictions before they are used as supervision targets. Finally, an overall loss function integrates supervised and unsupervised objectives, balancing cross-entropy on labeled data with consistency and rectification losses on unlabeled data. 

\subsubsection{\textit{Feature Enhancement}} 
Wavelet feature extraction plays a crucial role in leaf disease detection by capturing both spatial and frequency information from leaf images. The Discrete Wavelet Transform (DWT) decomposes an input image \( I \) into four distinct sub-band components: approximation (\( A \)) and details in the horizontal (\( H \)), vertical (\( V \)), and diagonal (\( D \)) directions. This decomposition is represented by~\autoref{DWT}.

\begin{equation}
\text{DWT}(I) = \{A, H, V, D\}
\label{DWT}
\end{equation}

The approximation component \( A \) captures the low-frequency structure of the image. At the same time, \( H \), \( V \), and \( D \) represent high-frequency components that preserve fine-grained textures and edge details related to disease symptoms, as described in \autoref{eq:9a}--\autoref{eq:9d}. These components are computed using convolution operations with wavelet filters, followed by downsampling:

\begin{subequations}
\label{AHVD}
\begin{align}
A &= (I * g_L^x * g_L^y)\downarrow 2, \label{eq:9a} \\
H &= (I * g_H^x * g_L^y)\downarrow 2, \label{eq:9b} \\
V &= (I * g_L^x * g_H^y)\downarrow 2, \label{eq:9c} \\
D &= (I * g_H^x * g_H^y)\downarrow 2. \label{eq:9d}
\end{align}
\end{subequations}

here, \( g_L \) and \( g_H \) are the low-pass and high-pass filters, and \( \downarrow 2 \) denotes downsampling by a factor of 2.
The convolution operator is denoted by \( * \). After extraction, the wavelet components are flattened and concatenated into a single feature vector, as shown in~\autoref{SFV}.

\begin{figure}[h!]
    \centering
    \includegraphics[width=\textwidth]{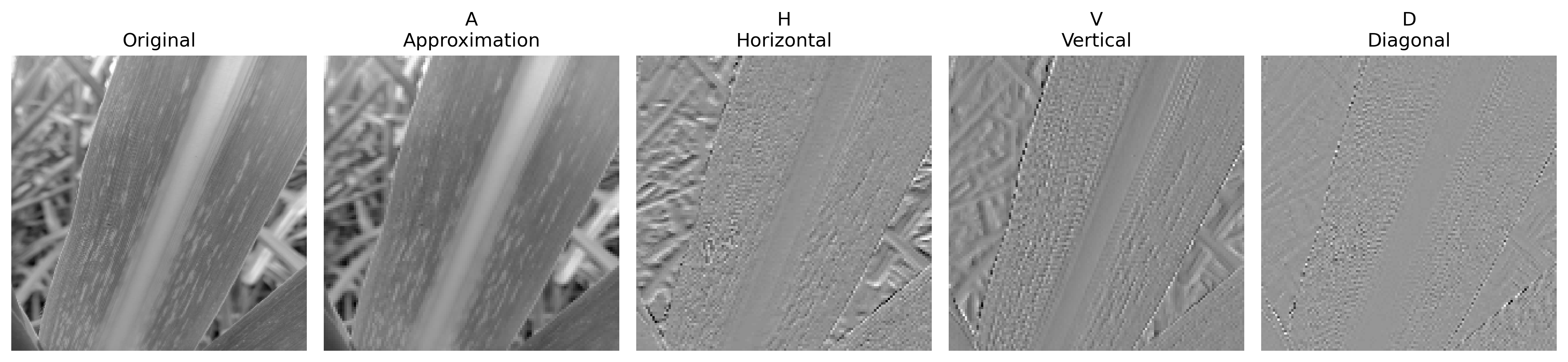} 
    \caption{DWT decomposition of a sugarcane leaf. Low-frequency (LL) captures the global structure; high-frequency sub-bands (LH, HL, HH) highlight textures and edges associated with lesions.}
    \label{fig:dwt}
\end{figure}

\begin{equation}
\mathbf{f}_{\text{wavelet}} = \text{concat}(\text{flatten}(A), \text{flatten}(H), \text{flatten}(V), \text{flatten}(D)) 
\label{SFV}
\end{equation}
{The approximation (A), horizontal (H), vertical (V), and diagonal (D) wavelet sub-bands are each transformed into one-dimensional vectors and subsequently combined to retain diverse information across multiple frequency scales, as shown in~\autoref{fig:dwt}. 
The approximation sub-band primarily captures the global shape and structural characteristics of the leaf, while the detail sub-bands emphasize directional patterns, texture, and edge information that are commonly associated with disease symptoms. 
Analysis of individual DWT sub-bands reveals that lesions with fine textures and small streaks produce strong responses in the high-frequency sub-bands (H, V, D). At the same time, overall leaf morphology is captured in the low-frequency sub-band (A). Visualizing these sub-bands demonstrates that DWT can highlight subtle disease patterns that are often missed by spatial-only CNN features, confirming that the transform targets specific frequency ranges associated with early-stage lesions.
The resulting concatenated wavelet feature vector provides a comprehensive frequency-domain representation, which is then processed and fused with features extracted using convolutional neural networks (CNNs). This fusion enables the model to automatically learn the relative importance of different frequency components during training. Consequently, the system achieves enhanced texture discrimination and more accurate identification of healthy and diseased leaf regions, particularly in the early stages of disease progression. Beyond class activation maps (CAMs), the DWT features provide complementary frequency-domain information that highlights subtle lesion patterns. High-frequency sub-bands (H, V, D) consistently emphasize small spots and streaks that are often missed by spatial-only representations, while the low-frequency sub-band (A) preserves global leaf structure. This additional frequency information improves the model's discriminative power, particularly for early-stage diseases, and complements CNN-based feature learning for more robust disease detection.}

\begin{figure*}[h!]
    \centering
    \includegraphics[width=12cm, height=7
    cm]{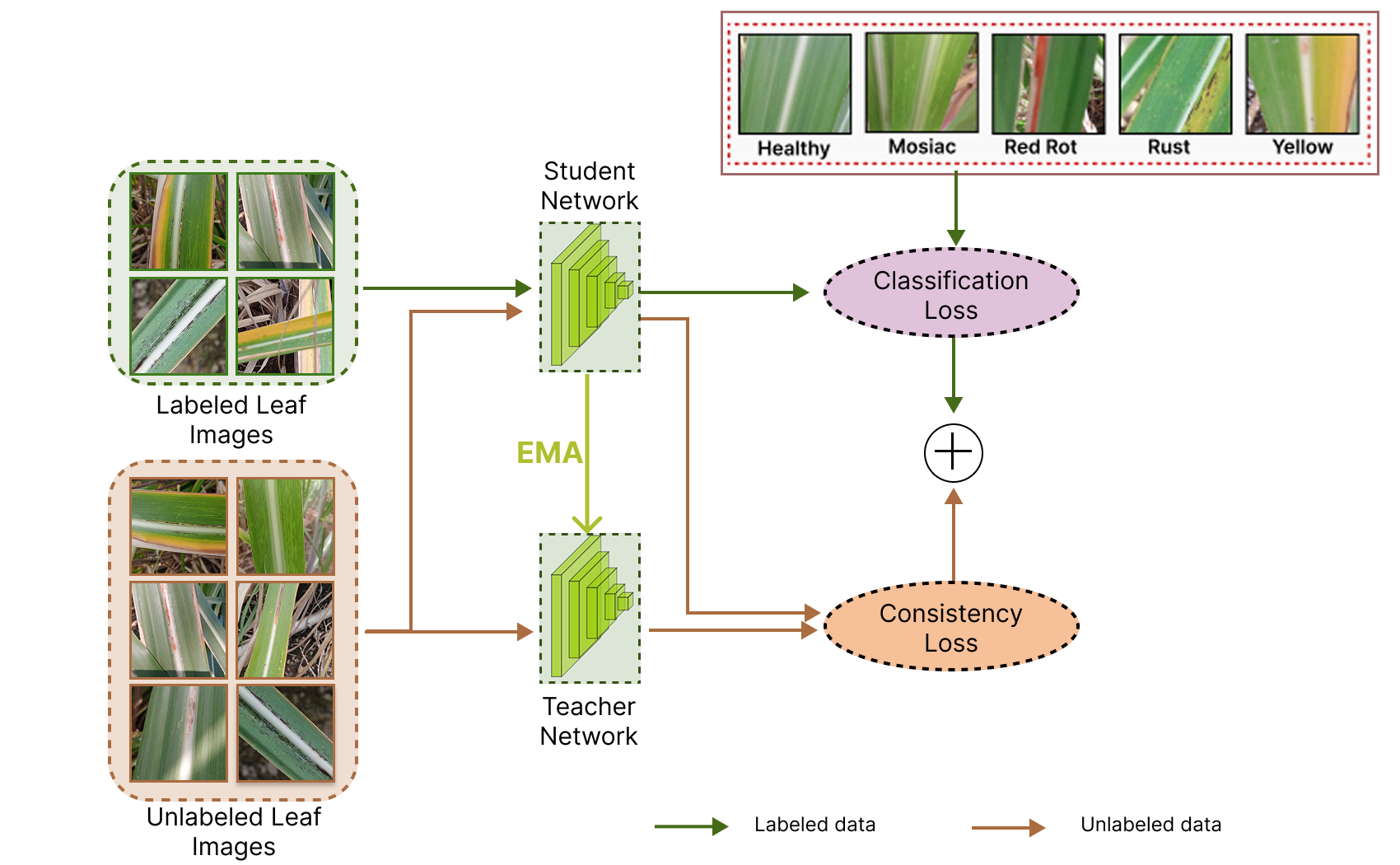}
    \caption{Basic Mean-Teacher model for sugarcane leaf disease detection.}
    \label{mean_teacher}
\end{figure*} 

\subsubsection{\textit{Mean-Teacher Model}} 
\begin{figure*}
    \centering
    \includegraphics[width=16cm, height=6cm]{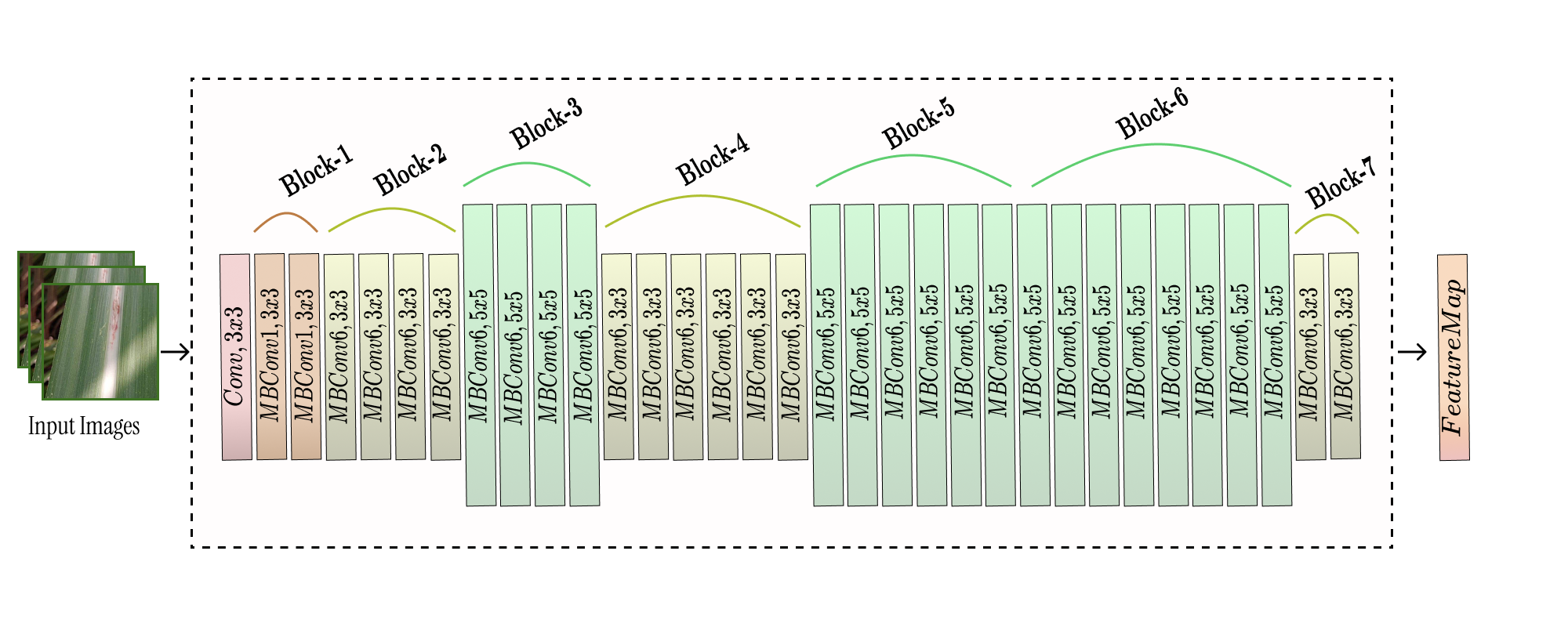}
    \caption{Architecture of EfficientNetB4 model.}
    \label{fig: EfficientNetB4}
\end{figure*}

The training dataset comprises both labeled and unlabeled images within the Mean-Teacher (MT) framework \citep{Tarvainen_2017}. Let \( x_i^l \) denote the \( i \)-th labeled image with its label \( y_i^l \), and \( x_j^u \) denote the \( j \)-th unlabeled image.
The MT method employs a unified setup between a student and a teacher model to improve learning from unlabeled data samples, as shown in~\autoref{mean_teacher}. Both models use the EfficientNetB4 architecture, which employs compound scaling of depth, width, and resolution to achieve high precision and efficiency, as shown in~\autoref{fig: EfficientNetB4}. 
The student model is trained on both labeled and unlabeled data, while the teacher model is updated using an Exponential Moving Average (EMA) of the student's weights. The teacher network generates pseudo-labels for unlabeled samples, while the student network aligns its predictions with the teacher's outputs to encourage stable and generalized training.

In this framework, both the student and teacher models follow the same architecture consisting of an encoder and a decoder. Their parameters are denoted by \( \theta_s \) for the student and \( \theta_t \) for the teacher. Importantly, only the student model is optimized through gradient descent, whereas the teacher model is updated using an Exponential Moving Average (EMA) and does not receive gradient updates. At iteration \( k \), the teacher parameters are updated as:  

\begin{equation}
  \theta_t^{(k)} = \lambda \cdot \theta_t^{(k-1)} + (1 - \lambda) \cdot \theta_s^{(k)} 
  \label{TP}
\end{equation}
where, \( \lambda \) is a smoothing factor, typically set to 0.99, controlling the balance between historical and recent information, as shown in~\autoref{TP}. In this work, \( \lambda \) is fixed at 0.99 throughout training, and this EMA formulation is used consistently across the framework. Labeled data is used to train the student in a supervised manner, while the teacher provides pseudo-labels for the unlabeled data to guide training.
{From a computational perspective, the proposed method introduces minimal overhead compared to standard supervised training. With a fixed input resolution, batch size, and training configuration, each training sample requires a constant amount of computation, including forward and backward passes through the student network and a forward pass through the teacher network with EMA-based updates. Therefore, the overall time complexity scales linearly with the number of training samples. Similarly, memory usage increases linearly with the dataset size due to input storage and intermediate feature representations, while the model parameters and EMA-related variables remain constant throughout training.}

\subsubsection{\textit{Pseudo Label Rectification}}
 
{Pseudo-label rectification addresses the issue of unreliable pseudo-labels by using consistency learning, which encourages the student model to produce predictions that are consistent with those of the teacher model \citep{Lu_2023}.}
The student generates predictions for both labeled and unlabeled samples and learns to match the teacher's stable outputs, computed using an Exponential Moving Average (EMA) of the student's weights. This process allows the student to learn from the teacher's guidance and improves the robustness to data deviations. The mutual interaction between the student and teacher also supports feature representation and improves the classification performance.
The consistency loss $\mathcal{L}_{\text{cons}}$, defined in~\autoref{ConsFinal}, incorporates both pointwise and distribution-level alignment.

\begin{equation}
\mathcal{L}_{\text{cons}} = \mathbb{E}_{x_u \sim \mathcal{D}_u} \Big[ \ell \big( f_{\text{student}}(x_u), f_{\text{teacher}}(x_u) \big) \Big] 
+ \lambda \sum_{i} p_{\text{teacher}}(x_i) \log \frac{p_{\text{teacher}}(x_i)}{p_{\text{student}}(x_i)},
\label{ConsFinal}
\end{equation}

Here, $\mathbb{E}_{x_u \sim \mathcal{D}_u}[\cdot]$ denotes the expectation over unlabeled samples $x_u$. {The functions $f_{\text{student}}(x_u)$ and $f_{\text{teacher}}(x_u)$ represent the softmax probability outputs of the student and teacher models, respectively. The first term measures the pointwise difference between predictions using the $L_2$ loss, while the second term represents the Kullback--Leibler (KL) divergence between the corresponding probability distributions. Unlike conventional formulations, the KL-divergence is not treated as an independent loss term but is used to estimate prediction uncertainty and guide the consistency learning process.} The weighting factor $\lambda$, typically set between $0.1$ and $1.0$, controls the contribution of the KL divergence term. Minimizing $\mathcal{L}_{\text{cons}}$ encourages the student model to align with the teacher model at both the prediction and distribution levels, thereby improving consistency and reducing uncertainty in pseudo-labels.

Selecting an effective backbone is important for achieving high performance in semi-supervised learning, particularly for tasks such as leaf disease detection. ResNet shows strong performance in general image classification. However, it often fails to capture the fine details needed to distinguish between different leaf diseases. EfficientNet achieves a better balance between accuracy and computational efficiency by scaling network depth, width, and input resolution together. It also extracts detailed features from sugarcane leaf images, improving classification with fewer parameters and less computation.

\subsubsection{\textit{Total Loss}}

The overall training objective of the proposed framework combines supervised classification, consistency regularization, and supervised contrastive learning. The total loss used to optimize the student network is defined in ~\autoref{eq:total_loss} as:

\begin{equation}
\mathcal{L}_{\text{total}} = \mathcal{L}_{\text{sup}} + \alpha \, \mathcal{L}_{\text{cons}} + \beta \, \mathcal{L}_{\text{SCL}}
\label{eq:total_loss}
\end{equation}

where \( \mathcal{L}_{\text{sup}} \) denotes the supervised classification loss computed on labeled samples, \( \mathcal{L}_{\text{cons}} \) represents the consistency loss imposed by the Mean-Teacher framework on unlabeled data, and \( \mathcal{L}_{\text{SCL}} \) is the supervised contrastive loss that promotes inter-class discrimination in the embedding space. 
{The weighting factors \( \alpha \) and \( \beta \) control the relative contributions of the consistency and contrastive loss terms, respectively. This combined objective enables the model to learn robust, discriminative, and semantically meaningful feature representations.}

\section{Results}
\label{Results}
In this section, we describe the experimental setup, the datasets used, the evaluation metrics, the qualitative results, and the ablation studies.

\subsection{Dataset description}
\begin{figure}[h!]
    \centering
    \begin{minipage}{0.245\textwidth}
        \centering
        \includegraphics[width=\textwidth]{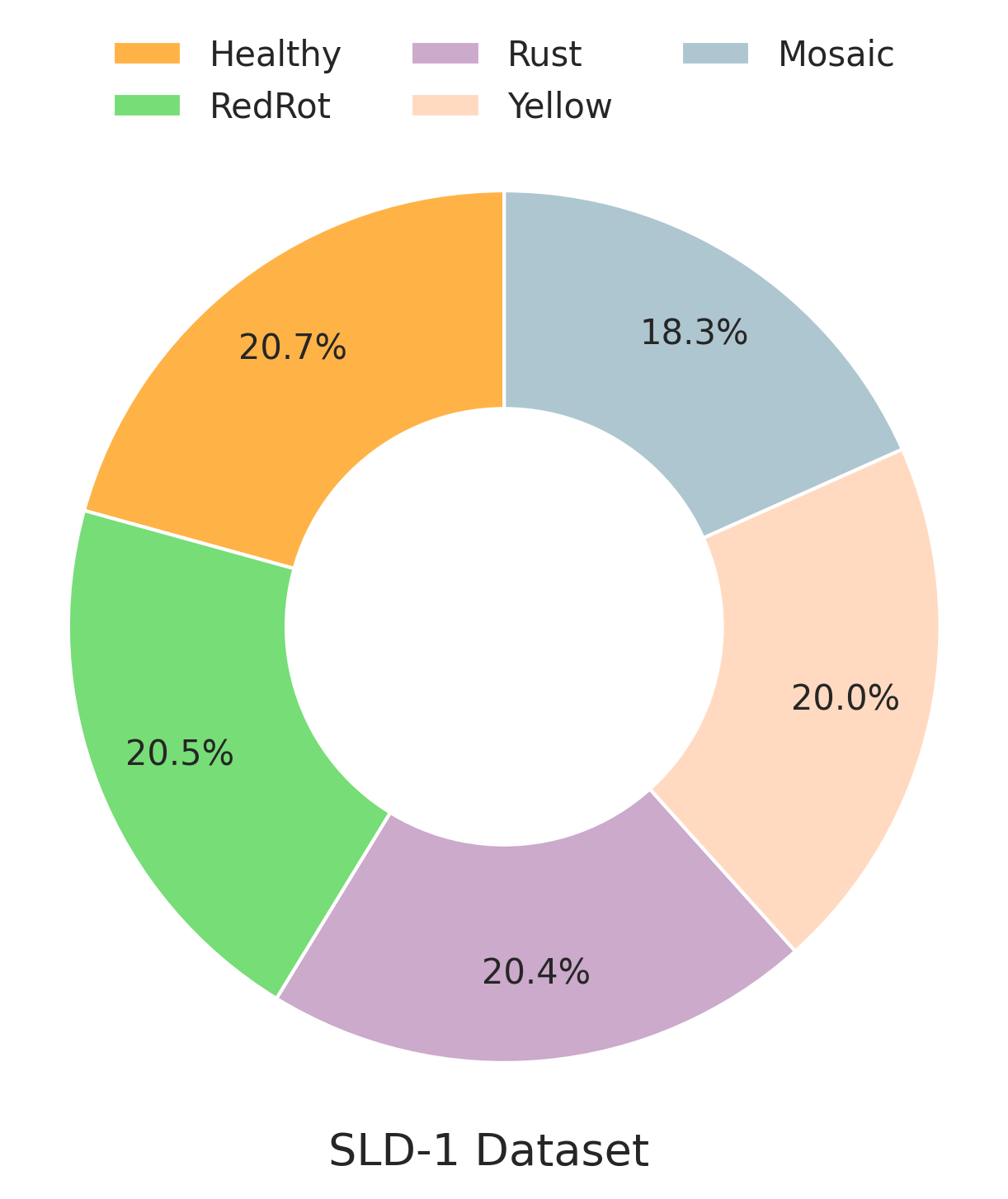}
    \end{minipage}%
    \hspace{0.0125cm}
    \begin{minipage}{0.3\textwidth}
        \centering
        \includegraphics[width=\textwidth]{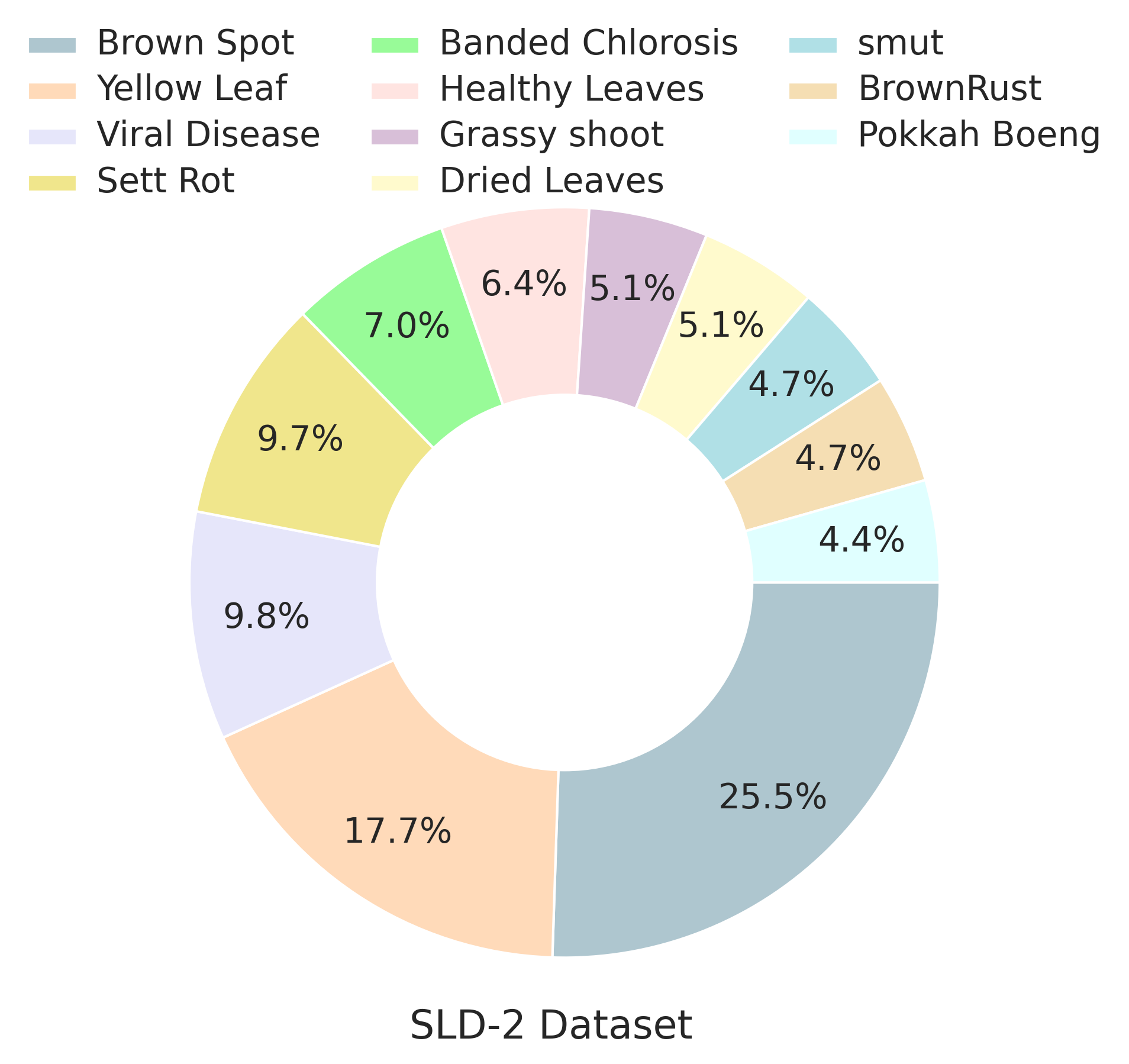}
    \end{minipage}%
    \hspace{0.0125cm}
    \begin{minipage}{0.229
        \textwidth}
        \centering
        \includegraphics[width=\textwidth]{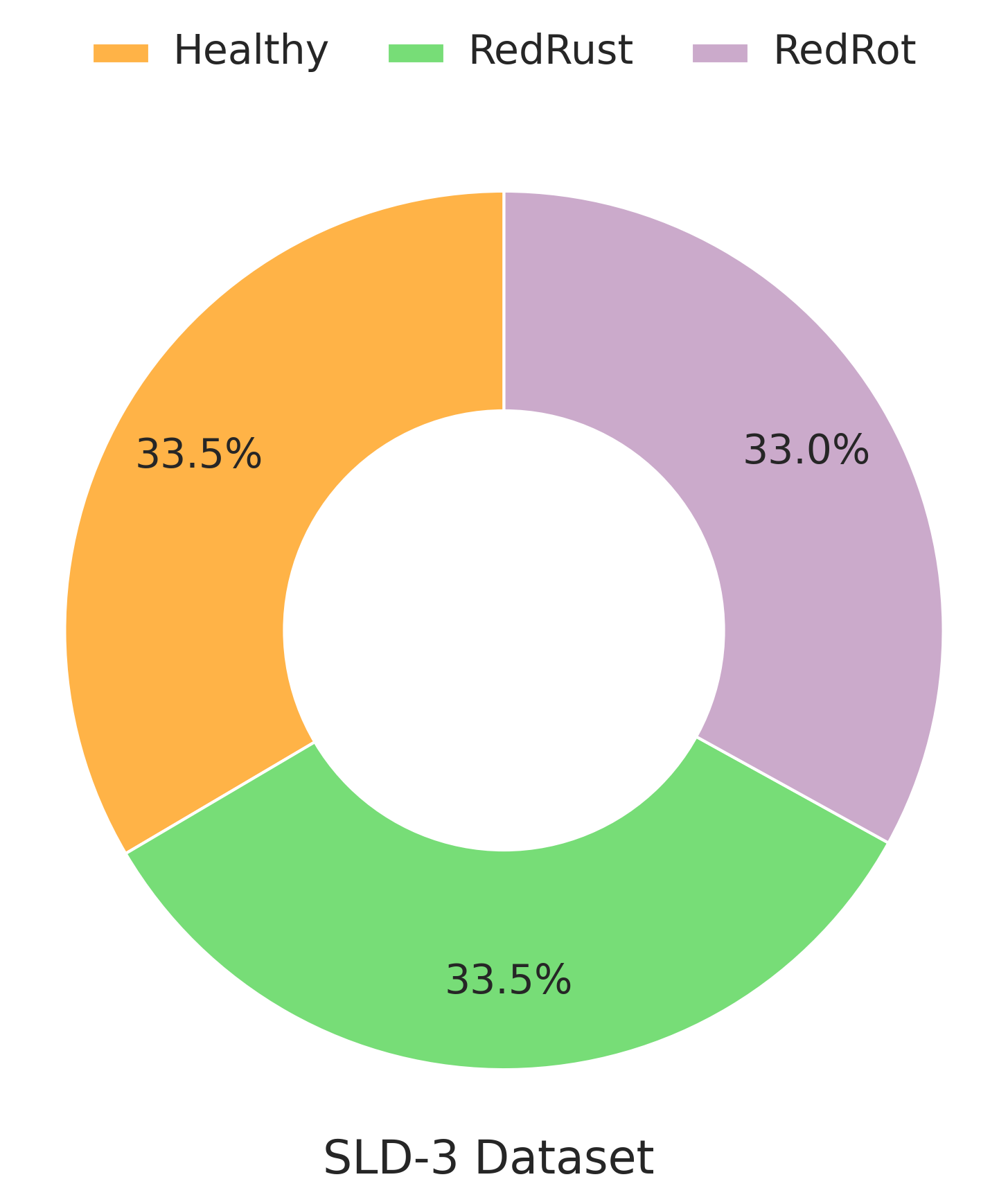}
    \end{minipage}
    \caption{Distribution of samples across various categories in SLD-1, SLD-2, and SLD-3.}
  \label{fig: pfc}
\end{figure}

The three publicly available datasets used in our study are shown in~\autoref{fig: pfc}. 

\begin{figure*}[h!]
    \centering
    \includegraphics[width=16cm, height=2.5cm]{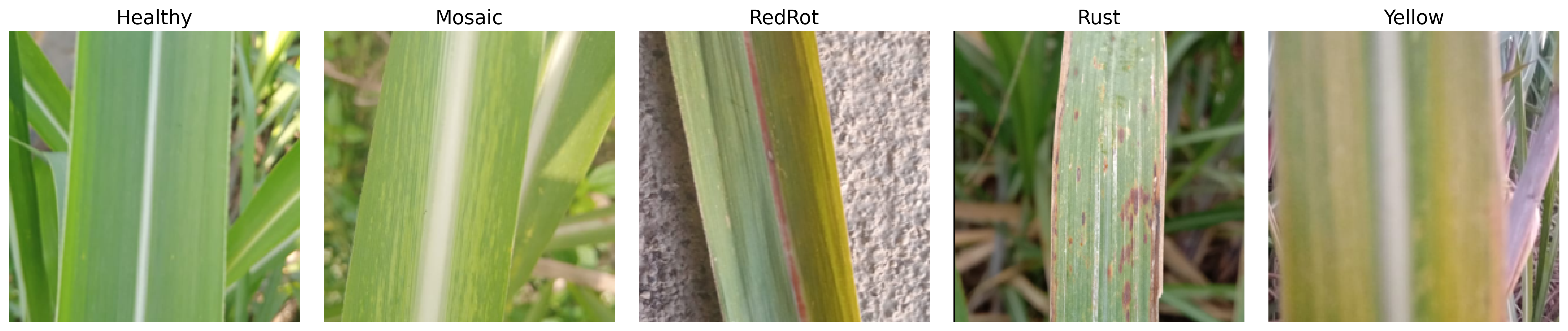}
    \caption{Illustration of sugarcane leaf disease types on the SLD-1 dataset.}
    \label{fig:SLD-1_Dataset}
\end{figure*}

\begin{figure*}[h!]
    \centering
    \includegraphics[width=16cm, height=7.5cm]{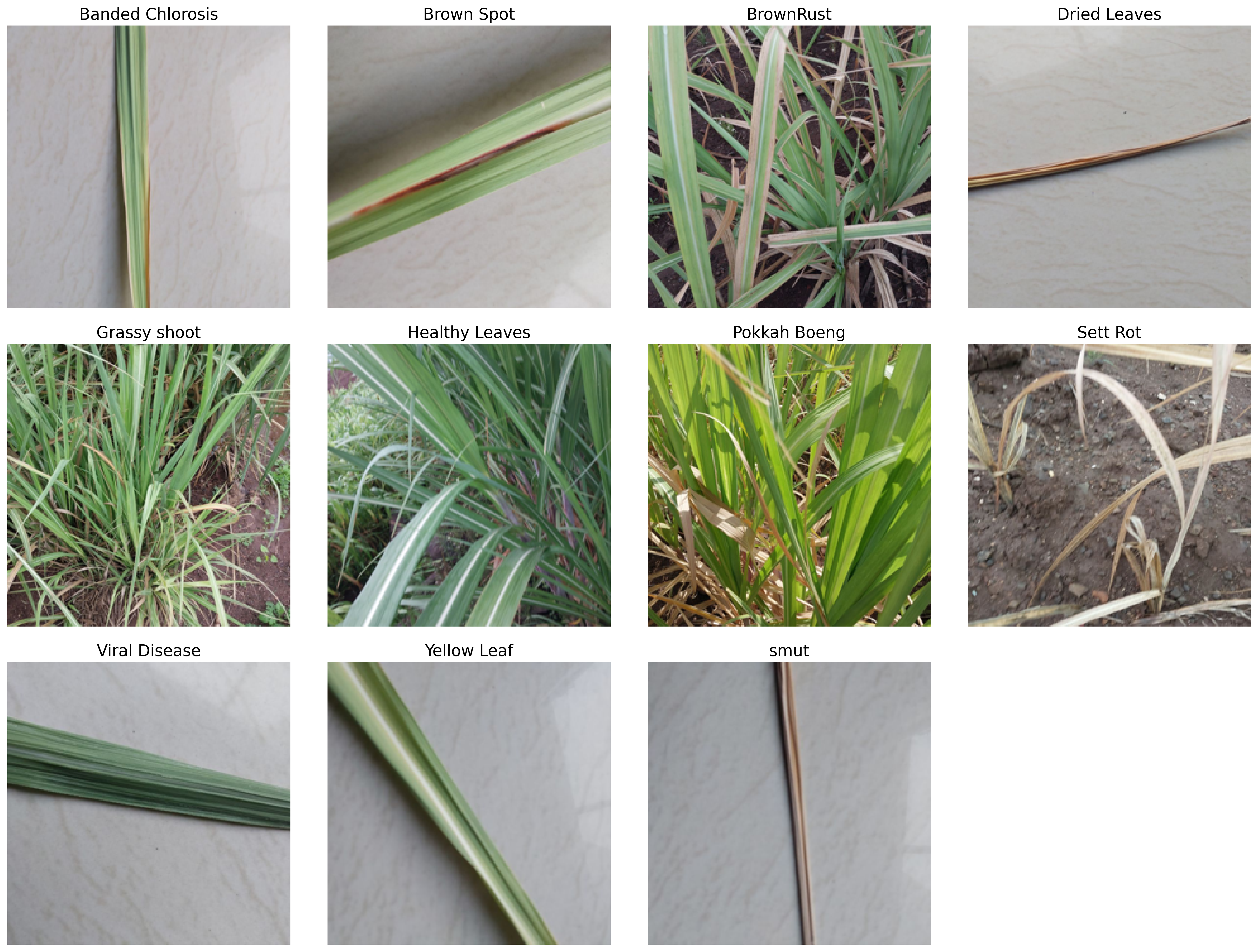}
    \caption{Overview of the SLD-2 dataset illustrating various sugarcane leaf disease types.}
    \label{fig:SLD-2_Dataset}
\end{figure*}

\begin{figure*}[h!]
    \centering
    \includegraphics[width=11cm, height=2.5cm]{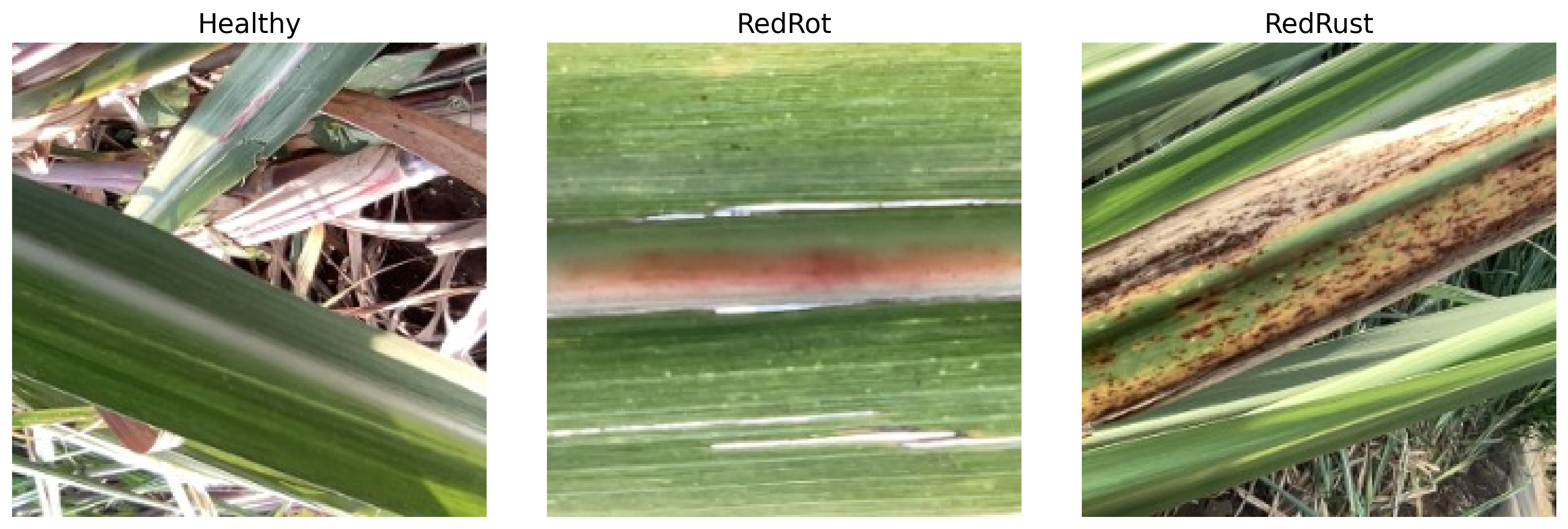}
    \caption{Overview of sugarcane leaf disease types on the SLD-3 dataset.}
    \label{fig:SLD-3_Dataset}
\end{figure*}

\subsubsection{SLD-1}
This SLD-1 sugarcane dataset consists of 2,521 images of sugarcane leaves collected from farms in Maharashtra, India \citep{Awais_2025, Anika_2025, Singh_2024, Paramanandham_2024, Nomani_2024}.
{The images were captured using smartphones with different camera specifications, which introduced variation in original image resolution, lighting conditions, and viewing angles. The SLD-1 dataset contains images with varying native resolutions.}
This dataset describes five categories: Healthy, Mosaic, Redrot, Rust, and Yellow, as described in ~\autoref{fig:SLD-1_Dataset}. All images are in RGB format, and the dataset is balanced across the five classes and provides a good level of variation within and between classes. This makes it suitable for training classification models in real-world conditions.

\subsubsection{SLD-2}
SLD-2 dataset \citep{Thite_2024} is the largest and comprises 6,748 high-resolution images of sugarcane leaves. The images are in JPEG format with a resolution of 764~$\times$~1024 pixels. The dataset covers 12 categories: 10 disease types, 1 healthy leaf class, and 1 dried leaf class, as shown in~\autoref{fig:SLD-2_Dataset}. The images were collected through extensive field surveys and captured from multiple angles and at different disease stages to reflect natural variations. High image quality and wide visual diversity make SLD-2 well-suited for fine-grained disease classification.

\subsubsection{SLD-3}
SLD-3 dataset \citep{Lambor_2022} is smaller and contains 224 RGB images organized into three classes: Healthy, Redrot, and Redrust, as defined in~\autoref{fig:SLD-3_Dataset}. The images were collected from sugarcane fields using the Google Teachable Machine platform, enabling quick image acquisition and labeling.
{All images in SLD-3 dataset have a fixed original resolution of 224 × 224 pixels.}

\subsection{Experimental Setup}
This section provides an overview of comparative methods. A summary of the hyperparameter configurations used in our proposed approach is presented in~\autoref{tab:hyperparameters}.
{For experimental evaluation, the dataset is divided using a fixed train--test split. In this setup, 10\% of the dataset is reserved as an independent test set, while the remaining 90\% is used for training the models. The test set is kept strictly unseen during training and used only for final performance evaluation to ensure an unbiased assessment of the model's generalization ability.}

\begin{table*}[htbp]\small
\centering
\caption{This table shows the final hyperparameter details of U-SSDNet.}
\label{tab:hyperparameters}
\begin{tabular*}{\textwidth}{@{\extracolsep{\fill}} l c}
\toprule
\textbf{Hyperparameter} & \textbf{Value} \\
\midrule
Image Size & 224$\times$224 \\
Batch Size & 8 \\
Optimizer & AdamW \\
Learning Rate & 0.001 \\
Number of Epochs & 35 \\
Weight Decay & 0.0001 \\
Activation Function & ReLU \\
\bottomrule
\end{tabular*}
\end{table*}

\subsubsection{Comparison Methods}
The various state-of-the-art methods used in our study are discussed as: MAIAnet \citep{Zhang_2024},  SoyaTrans \citep{Sharma_2025}, ResTS \citep{Shah_2022}, TSwin-F \citep{Ye_2024}, ConvNext \citep{Benchallal_2024}, and CorePlant \citep{Ilsever_2024}.

\vspace{2mm}
 \textbf{MAIAnet} \citep{Zhang_2024} combines attention mechanisms with normalization layers to improve the precision of cassava leaf disease detection.
 
\vspace{2mm}
 \textbf{SoyaTrans} \citep{Sharma_2025} propose a transformer-based model for leaf disease classification, focusing on fine-grained feature extraction in challenging real-world environments.
 
\vspace{2mm}
 \textbf{ResTS} \citep{Shah_2022} introduces a residual deep learning approach for detecting plant diseases and improves both model interpretability and performance by focusing on the informative leaf regions.
 
\vspace{2mm}
 \textbf{TSwin-F} \citep{Ye_2024} adopts a transformer model with multi-scale feature fusion to identify tomato leaf infections and to capture local characteristics as well as broader spatial relationships.
 
\vspace{2mm}
 \textbf{ConvNext} \citep{Benchallal_2024} applies semi-supervised consistency regularization to achieve reliable classification results, even when only a small set of labeled samples is available.
 
\vspace{2mm}
\textbf{CorePlant} \citep{Ilsever_2024} uses consistency regularization to combine labeled and unlabeled data, reducing labeling effort while maintaining performance.

\subsubsection{Evaluation Metrics for Leaf Disease Classification}
To assess the performance of our method, we use four standard evaluation metrics: Accuracy (Acc), Precision (Pre), Recall (Rec), and F1-score (F1).

\subsection{Experimental Results}
We present qualitative results comparing state-of-the-art methods with our proposed approach. We also perform ablation analysis to investigate the contribution and effectiveness of individual components in the proposed framework.

\subsubsection{Comparative Analysis with State-of-the-art Methods}
We compare the performance of our proposed method with state-of-the-art methods on three datasets.
\paragraph{\textbf{(a) On SLD-1 Dataset}}~\autoref{tab:Method_SLD1_mean_std} presents the performance of our proposed U-SSDNet approach compared to state-of-the-art leaf disease classification methods across varying labeled data proportions (10--50\%). At all data ratios, U-SSDNet consistently outperforms both supervised and semi-supervised baselines, achieving the highest accuracy and F1-scores, for example, 94.80\% and 94.20\% at 10\% labeled data, and up to 97.83\% and 97.42\% at 50\% labeled data. Among supervised methods, CorePlant \citep{Ilsever_2024} shows strong performance at higher labeled data ratios but remains below the proposed approach, while ResTS \citep{Shah_2022} and SoyaTrans \citep{Sharma_2025} improve moderately with more labeled data but lag at lower data proportions. MAIAnet \citep{Zhang_2024} and TSwin-F \citep{Ye_2024} capture leaf disease patterns but achieve only moderate performance, potentially struggling
\begin{table}[h!]
\centering
\setlength{\tabcolsep}{4pt}
\renewcommand{\arraystretch}{1.4}
\caption{Quantitative evaluation of supervised and semi-supervised approaches on the SLD-1 dataset using mean $\pm$ standard deviation.}
\label{tab:Method_SLD1_mean_std}
\begin{adjustbox}{max width=\textwidth}
\begin{tabular}{|c|c|cc|cc|cc|cc|cc|}
\hline
\multirow{2}{*}{\textbf{Approach}} & \multirow{2}{*}{\textbf{Model}} 
& \multicolumn{2}{c|}{\textbf{10\%}} 
& \multicolumn{2}{c|}{\textbf{20\%}} 
& \multicolumn{2}{c|}{\textbf{30\%}} 
& \multicolumn{2}{c|}{\textbf{40\%}} 
& \multicolumn{2}{c|}{\textbf{50\%}} \\
\cline{3-12}me
& & \textbf{Acc} & \textbf{F1} 
& \textbf{Acc} & \textbf{F1} 
& \textbf{Acc} & \textbf{F1} 
& \textbf{Acc} & \textbf{F1} 
& \textbf{Acc} & \textbf{F1} \\
\hline

\multirow{4}{*}{Supervised} 

& MAIAnet \citep{Zhang_2024} 
& $35.76 \pm 1.98$ & $23.12 \pm 1.19$ 
& $40.14 \pm 2.12$ & $26.62 \pm 1.22$ 
& $47.08 \pm 1.63$ & $42.53 \pm 1.34$ 
& $53.15 \pm 1.57$ & $49.96 \pm 1.68$ 
& $54.96 \pm 1.89$ & $50.96 \pm 1.02$ \\

& SoyaTrans \citep{Sharma_2025} 
& $56.18 \pm 1.66$ & $54.69 \pm 1.81$ 
& $70.31 \pm 1.57$ & $69.38 \pm 1.56$ 
& $72.61 \pm 1.90$ & $69.45 \pm 1.82$ 
& $76.78 \pm 2.01$ & $75.91 \pm 1.61$ 
& $79.55 \pm 2.72$ & $78.14 \pm 2.89$ \\

& ResTS \citep{Shah_2022} 
& $43.41 \pm 1.85$ & $38.23 \pm 0.92$ 
& $59.32 \pm 0.64$ & $56.39 \pm 1.03$ 
& $64.91 \pm 1.57$ & $63.01 \pm 1.62$ 
& $76.02 \pm 1.85$ & $74.94 \pm 1.21$ 
& $83.88 \pm 1.38$ & $82.03 \pm 1.68$ \\

& TSwin-F \citep{Ye_2024} 
& $20.74 \pm 1.86$ & $12.42 \pm 2.15$ 
& $29.65 \pm 1.43$ & $22.82 \pm 1.59$ 
& $32.74 \pm 1.71$ & $23.61 \pm 1.59$ 
& $34.16 \pm 1.27$ & $27.37 \pm 1.15$ 
& $34.85 \pm 1.72$ & $28.93 \pm 1.80$ \\

\hline

\multirow{2}{*}{Semi-supervised} 

& ConvNext \citep{Benchallal_2024} 
& $49.37 \pm 1.78$ & $47.49 \pm 1.05$ 
& $55.20 \pm 1.94$ & $54.48 \pm 1.98$ 
& $61.28 \pm 1.74$ & $60.44 \pm 0.83$ 
& $64.89 \pm 1.83$ & $64.45 \pm 0.90$ 
& $69.14 \pm 1.29$ & $68.81 \pm 0.80$ \\

& CorePlant \citep{Ilsever_2024} 
& $64.82 \pm 2.46$ & $58.02 \pm 2.39$ 
& $78.95 \pm 1.89$ & $77.80 \pm 1.73$ 
& $85.17 \pm 2.41$ & $84.90 \pm 2.39$ 
& $87.49 \pm 1.82$ & $86.37 \pm 1.86$ 
& $88.72 \pm 1.96$ & $87.32 \pm 2.04$ \\

\hline

\multirow{1}{*}{Proposed} 

& \textbf{U-SSDNet} 
& $\mathbf{94.80 \pm 1.40}$ & $\mathbf{94.20 \pm 1.60}$ 
& $\mathbf{95.62 \pm 0.49}$ & $\mathbf{95.61 \pm 0.57}$ 
& $\mathbf{96.82 \pm 0.50}$ & $\mathbf{96.37 \pm 0.47}$ 
& $\mathbf{97.17 \pm 1.07}$ & $\mathbf{96.38 \pm 1.38}$ 
& $\mathbf{97.83 \pm 1.27}$ & $\mathbf{97.42 \pm 1.36}$ \\

\hline
\end{tabular}
\end{adjustbox}
\end{table}
with subtle or complex disease features. The semi-supervised ConvNext approach \citep{Benchallal_2024} leverages unlabeled data, thereby improving its ability to recognize disease patterns compared to fully supervised methods. The superior performance of U-SSDNet stems from its hybrid design, which integrates global context modeling with local feature extraction to capture both long-range dependencies and complex leaf-disease patterns.
 
\paragraph{\textbf{(b) On SLD-2 Dataset}}
For different labeled data proportions ranging from 10\% to 50\%,~\autoref{tab:Method_SLD2_mean_std_2} presents quantitative results for the U-SSDNet method on the SLD-2 dataset. The U-SSDNet method consistently performs better in terms of accuracy and F1-score across all labeled data ratios, ranging from 90.87\% and 89.88\% at 10\% labeled data to 95.07\% and 94.73\% at 50\% labeled data. CorePlant \citep{Ilsever_2024} performs strongly at higher data proportions but lags behind U-SSDNet at lower labeled data. ResTS \citep{Shah_2022} and SoyaTrans \citep{Sharma_2025} show slow improvement with more training samples and perform worse than the proposed method when there is less labeled data. MAIAnet \citep{Zhang_2024} and TSwin-F \citep{Ye_2024} gain average results and indicate challenges in learning fine disease patterns. Semi-supervised ConvNext \citep{Benchallal_2024} benefits from unlabeled data, but its performance remains below U-SSDNet across all splits. The performance of U-SSDNet is better at capturing complex disease patterns and subtle visual cues, and at robust classification across all leaf disease classes.

\begin{table}[h!]
\centering
\setlength{\tabcolsep}{4pt}
\renewcommand{\arraystretch}{1.4}
\caption{Quantitative evaluation of supervised and semi-supervised approaches on the SLD-2 dataset using mean $\pm$ standard deviation.}
\label{tab:Method_SLD2_mean_std_2}
\begin{adjustbox}{max width=\textwidth}
\begin{tabular}{|c|c|cc|cc|cc|cc|cc|}
\hline
\multirow{2}{*}{\textbf{Approach}} & \multirow{2}{*}{\textbf{Model}} 
& \multicolumn{2}{c|}{\textbf{10\%}} 
& \multicolumn{2}{c|}{\textbf{20\%}} 
& \multicolumn{2}{c|}{\textbf{30\%}} 
& \multicolumn{2}{c|}{\textbf{40\%}} 
& \multicolumn{2}{c|}{\textbf{50\%}} \\
\cline{3-12}
& & \textbf{Acc} & \textbf{F1} 
& \textbf{Acc} & \textbf{F1} 
& \textbf{Acc} & \textbf{F1} 
& \textbf{Acc} & \textbf{F1} 
& \textbf{Acc} & \textbf{F1} \\
\hline

\multirow{4}{*}{Supervised} 

& MAIAnet \citep{Zhang_2024} 
& $29.56 \pm 2.35$ & $25.76 \pm 1.81$ 
& $30.34 \pm 1.75$ & $27.35 \pm 1.68$ 
& $35.69 \pm 1.83$ & $28.57 \pm 2.15$ 
& $38.54 \pm 1.28$ & $30.27 \pm 1.83$ 
& $39.73 \pm 1.85$ & $31.35 \pm 1.74$ \\

& SoyaTrans \citep{Sharma_2025} 
& $47.25 \pm 1.71$ & $35.82 \pm 1.93$ 
& $58.03 \pm 1.57$ & $50.22 \pm 1.40$ 
& $63.25 \pm 1.78$ & $58.02 \pm 1.61$ 
& $66.94 \pm 1.35$ & $61.44 \pm 1.42$ 
& $72.64 \pm 1.71$ & $69.73 \pm 1.80$ \\

& ResTS \citep{Shah_2022} 
& $48.19 \pm 0.69$ & $43.25 \pm 0.75$ 
& $60.81 \pm 1.37$ & $56.98 \pm 1.58$ 
& $72.02 \pm 0.81$ & $70.26 \pm 0.79$ 
& $80.74 \pm 1.51$ & $78.73 \pm 1.40$ 
& $86.78 \pm 1.93$ & $85.91 \pm 1.78$ \\

& TSwin-F \citep{Ye_2024} 
& $26.82 \pm 1.90$ & $16.74 \pm 1.96$ 
& $34.16 \pm 0.92$ & $25.14 \pm 0.87$ 
& $36.87 \pm 1.31$ & $25.85 \pm 1.60$ 
& $37.37 \pm 0.79$ & $26.18 \pm 1.36$ 
& $38.19 \pm 2.31$ & $28.30 \pm 1.81$ \\

\hline

\multirow{2}{*}{Semi-supervised} 

& ConvNext \citep{Benchallal_2024} 
& $63.18 \pm 1.94$ & $59.62 \pm 1.80$ 
& $69.46 \pm 1.59$ & $62.79 \pm 1.67$ 
& $75.02 \pm 1.47$ & $72.90 \pm 1.41$ 
& $79.51 \pm 0.93$ & $73.93 \pm 1.13$ 
& $83.97 \pm 1.81$ & $81.08 \pm 1.85$ \\

& CorePlant \citep{Ilsever_2024} 
& $74.68 \pm 0.88$ & $66.15 \pm 0.82$ 
& $83.72 \pm 1.31$ & $79.35 \pm 1.28$ 
& $90.22 \pm 1.67$ & $88.02 \pm 1.58$ 
& $90.74 \pm 1.35$ & $88.13 \pm 1.50$ 
& $90.87 \pm 1.17$ & $88.64 \pm 1.26$ \\

\hline

\multirow{1}{*}{Proposed} 

& \textbf{U-SSDNet} 
& $\mathbf{90.87 \pm 1.87}$ & $\mathbf{89.88 \pm 3.51}$ 
& $\mathbf{92.85 \pm 0.55}$ & $\mathbf{92.95 \pm 1.20}$ 
& $\mathbf{93.05 \pm 1.38}$ & $\mathbf{92.74 \pm 1.82}$ 
& $\mathbf{93.79 \pm 0.65}$ & $\mathbf{93.46 \pm 1.77}$ 
& $\mathbf{95.07 \pm 0.32}$ & $\mathbf{94.73 \pm 0.30}$ \\

\hline
\end{tabular}
\end{adjustbox}
\end{table}

\paragraph{\textbf{(c) On SLD-3 Dataset}}
~\autoref{tab:Method_SLD3_mean_std_3} shows the relative performance of the proposed U-SSDNet framework on the SLD-3 dataset against existing supervised and semi-supervised methods. Supervised methods, such as SoyaTrans \citep{Sharma_2025} and ResTS \citep{Shah_2022}, show continuous improvements with larger proportions of labeled data but remain below U-SSDNet. MAIAnet \citep{Zhang_2024} and TSwin-F \citep{Ye_2024} achieve lower scores and struggle to capture fine and complex disease patterns.  
Semi-supervised models such as ConvNext (Benchallal, 2024) and CorePlant (Ilsever, 2024) perform better with more labeled data, but they still do not match the accuracy and consistency of U-SSDNet. The U-SSDNet shows superior and stable performance for all labeled proportions, reaching up to 98.55\% accuracy and 98.91\% F1-score, and confirming its robustness for reliable leaf disease classification on the SLD-3 dataset.

\begin{table}[h!]
\centering
\setlength{\tabcolsep}{4pt}
\renewcommand{\arraystretch}{1.4}
\caption{Quantitative evaluation of supervised and semi-supervised approaches on the SLD-3 dataset using mean $\pm$ standard deviation.}
\label{tab:Method_SLD3_mean_std_3}
\begin{adjustbox}{max width=\textwidth}
\begin{tabular}{|c|c|cc|cc|cc|cc|cc|}
\hline
\multirow{2}{*}{\textbf{Approach}} & \multirow{2}{*}{\textbf{Model}} 
& \multicolumn{2}{c|}{\textbf{10\%}} 
& \multicolumn{2}{c|}{\textbf{20\%}} 
& \multicolumn{2}{c|}{\textbf{30\%}} 
& \multicolumn{2}{c|}{\textbf{40\%}} 
& \multicolumn{2}{c|}{\textbf{50\%}} \\
\cline{3-12}
& & \textbf{Acc} & \textbf{F1} 
& \textbf{Acc} & \textbf{F1} 
& \textbf{Acc} & \textbf{F1} 
& \textbf{Acc} & \textbf{F1} 
& \textbf{Acc} & \textbf{F1} \\
\hline

\multirow{4}{*}{Supervised} 

& MAIAnet \citep{Zhang_2024} 
& $37.33 \pm 1.76$ & $34.88 \pm 1.50$ 
& $42.47 \pm 2.85$ & $40.78 \pm 2.54$ 
& $51.52 \pm 1.21$ & $48.46 \pm 1.22$ 
& $52.48 \pm 1.53$ & $50.63 \pm 1.48$ 
& $55.15 \pm 1.64$ & $53.20 \pm 1.58$ \\

& SoyaTrans \citep{Sharma_2025} 
& $52.64 \pm 1.49$ & $48.83 \pm 1.40$ 
& $59.61 \pm 1.37$ & $55.78 \pm 1.42$ 
& $66.89 \pm 1.87$ & $65.02 \pm 1.89$ 
& $73.71 \pm 1.05$ & $69.92 \pm 1.17$ 
& $78.06 \pm 1.45$ & $76.25 \pm 1.44$ \\

& ResTS \citep{Shah_2022} 
& $42.08 \pm 1.87$ & $36.91 \pm 1.58$ 
& $48.59 \pm 0.81$ & $42.36 \pm 1.11$ 
& $52.71 \pm 1.63$ & $48.47 \pm 1.52$ 
& $57.21 \pm 1.47$ & $54.91 \pm 1.27$ 
& $64.35 \pm 1.38$ & $59.48 \pm 1.25$ \\

& TSwin-F \citep{Ye_2024} 
& $32.37 \pm 1.81$ & $20.13 \pm 2.09$ 
& $36.43 \pm 1.88$ & $24.32 \pm 1.60$ 
& $37.16 \pm 1.73$ & $26.01 \pm 1.59$ 
& $43.17 \pm 1.83$ & $30.14 \pm 1.57$ 
& $58.31 \pm 1.79$ & $50.63 \pm 2.08$ \\

\hline

\multirow{2}{*}{Semi-supervised} 

& ConvNext \citep{Benchallal_2024} 
& $61.78 \pm 1.04$ & $58.79 \pm 1.18$ 
& $65.94 \pm 1.80$ & $63.14 \pm 1.77$ 
& $70.83 \pm 1.69$ & $68.87 \pm 1.63$ 
& $89.21 \pm 1.35$ & $88.81 \pm 1.34$ 
& $92.07 \pm 1.25$ & $91.96 \pm 1.48$ \\

& CorePlant \citep{Ilsever_2024} 
& $62.50 \pm 1.91$ & $53.82 \pm 1.80$ 
& $89.91 \pm 1.38$ & $67.46 \pm 1.41$ 
& $90.38 \pm 0.77$ & $90.36 \pm 0.67$ 
& $92.73 \pm 1.22$ & $90.49 \pm 1.50$ 
& $92.88 \pm 1.03$ & $92.01 \pm 1.42$ \\

\hline

\multirow{1}{*}{Proposed} 

& \textbf{U-SSDNet} 
& $\mathbf{96.76 \pm 1.78}$ & $\mathbf{97.10 \pm 1.65}$ 
& $\mathbf{97.18 \pm 1.82}$ & $\mathbf{97.40 \pm 1.67}$ 
& $\mathbf{97.74 \pm 1.26}$ & $\mathbf{98.80 \pm 0.21}$ 
& $\mathbf{98.27 \pm 1.69}$ & $\mathbf{98.87 \pm 1.43}$ 
& $\mathbf{98.55 \pm 1.47}$ & $\mathbf{98.91 \pm 1.38}$ \\

\hline
\end{tabular}
\end{adjustbox}
\end{table}
\subsubsection{Comparative Analysis with the Baseline Models}
{\autoref{tab:model_sld-1} and \autoref{tab:model_sld-2} present the performance of the proposed U-SSDNet compared with state-of-the-art models on  SLD-1 and SLD-2 datasets under different labeled data ratios. On the SLD-1 dataset with 50\% labeled data, U-SSDNet achieves an F1-score of 97.42\%, outperforming classical CNNs such as ResNet-50 (93.67\%), DenseNet-121 (93.22\%), and VGG-16 (70.14\%), as well as transformer-based models including ViT (93.42\%) and Swin Transformer (85.66\%). These results indicate that U-SSDNet can effectively utilize the available labeled data to achieve high performance.}

When the amount of labeled data is reduced to 20\% and 10\%, U-SSDNet still maintains superior performance compared with both conventional networks and semi-supervised methods, including FixMatch, FlexMatch, SoftMatch, and GapMatch. As shown in Table~\ref{tab:model_sld-2} on the SLD-2 dataset, the proposed method achieves F1-scores of 94.73\%, 92.95\%, and 89.88\% for 50\%, 20\%, and 10\% labeled data, respectively. This demonstrates that U-SSDNet is robust in scenarios with limited labeled data and consistently outperforms existing approaches.

\begin{table}[h!]
\centering
\caption{Performance comparison of SOTA models and the proposed U-SSDNet framework on varying labeled data proportions on the SLD-1 dataset.}
\label{tab:model_sld-1}
\resizebox{\textwidth}{!}{%
\begin{tabular}{lcccccccccccc}
\toprule
\multirow{2}{*}{\textbf{Model}} & \multicolumn{4}{c}{\textbf{50\% Labeled}} & \multicolumn{4}{c}{\textbf{20\% Labeled}} & \multicolumn{4}{c}{\textbf{10\% Labeled}} \\
 & \textbf{Acc} & \textbf{Pre} & \textbf{Rec} & \textbf{F1} 
 & \textbf{Acc} & \textbf{Pre} & \textbf{Rec} & \textbf{F1}
 & \textbf{Acc} & \textbf{Pre} & \textbf{Rec} & \textbf{F1} \\
\midrule

ResNet-50 \citep{He_2016} & 93.67 & 93.77 & 93.67 & 93.67 & 87.72 & 87.87 & 87.72 & 87.52 & 82.97 & 84.31 & 82.97 & 83.03 \\
VGG-16 \citep{Liu_2015} & 70.50 & 70.45 & 70.50 & 70.14 & 53.07 & 54.27 & 53.07 & 52.88 & 46.93 & 48.82 & 46.93 & 47.03 \\
DenseNet-121 \citep{Huang_2017} & 93.27 & 93.33 & 93.27 & 93.22 & 90.30 & 90.22 & 90.30 & 90.21 & 70.20 & 70.22 & 71.02 & 70.22 \\
ViT \citep{Dosovitskiy_2020} & 93.42 & 94.22 & 93.42 & 93.42 & 70.22 & 71.82 & 70.71 & 70.81 & 50.02 & 49.05 & 50.04 & 50.55 \\
Swin Transformer \citep{Liu_2021} & 85.66 & 85.45 & 85.66 & 85.66 & 75.52 & 77.45 & 75.52 & 75.66 & 45.45 & 45.67 & 45.45 & 45.49 \\
Mean Teacher \citep{Tarvainen_2017} & 91.52 & 92.55 & 91.52 & 91.98 & 89.78 & 89.90 & 89.78 & 89.92 & 87.22 & 88.32 & 87.22 & 87.29 \\
FixMatch \citep{Sohn_2020_FixMatch} & 97.25 & 97.46 & 97.14 & 97.22 & 89.80 & 89.80 & 89.72 & 89.69 & 87.45 & 87.45 & 87.32 & 87.33 \\
FlexMatch \citep{Zhang_2021_FlexMatch} & 89.02 & 90.43 & 89.16 & 88.89 & 88.63 & 90.10 & 88.81 & 88.60 & 89.80 & 90.47 & 89.57 & 89.72 \\
SoftMatch \citep{Chen_2023_SoftMatch} & 95.60 & 95.78 & 95.60 & 95.61 & 90.00 & 90.09 & 90.00 & 90.01 & 79.60 & 79.91 & 79.60 & 79.61 \\
GapMatch \citep{Huang_2025} & 93.45 & 93.70 & 93.14 & 92.89 & 91.90 & 90.67 & 91.90 & 90.13 & 90.19 & 90.58 & 90.19 & 90.01 \\
\midrule

\textbf{U-SSDNet (Proposed)} & \textbf{97.83} & \textbf{97.91} & \textbf{97.37} & \textbf{97.42}
& \textbf{95.62} & \textbf{95.38} & \textbf{95.88} & \textbf{95.61}
& \textbf{94.80} & \textbf{94.82} & \textbf{94.88} & \textbf{94.20} \\
\bottomrule
\end{tabular}%
}
\end{table}

\begin{table}[h!]
\centering
\caption{Performance comparison of SOTA models and the proposed U-SSDNet framework under different labeled data proportions on the SLD-2 dataset.}
\label{tab:model_sld-2}
\resizebox{\textwidth}{!}{%
\begin{tabular}{lcccccccccccc}
\toprule
\multirow{2}{*}{\textbf{Model}} & \multicolumn{4}{c}{\textbf{50\% Labeled}} & \multicolumn{4}{c}{\textbf{20\% Labeled}} & \multicolumn{4}{c}{\textbf{10\% Labeled}} \\
 & \textbf{Acc} & \textbf{Pre} & \textbf{Rec} & \textbf{F1}
 & \textbf{Acc} & \textbf{Pre} & \textbf{Rec} & \textbf{F1}
 & \textbf{Acc} & \textbf{Pre} & \textbf{Rec} & \textbf{F1} \\
\midrule

ResNet-50 \citep{He_2016} & 91.48 & 91.73 & 91.48 & 91.51 & 92.07 & 92.03 & 92.07 & 91.96 & 85.48 & 85.18 & 85.48 & 84.93 \\
VGG-16 \citep{Liu_2015} & 93.93 & 94.12 & 93.93 & 93.95 & 91.48 & 91.51 & 91.48 & 91.46 & 80.44 & 82.73 & 80.44 & 79.42 \\
DenseNet-121 \citep{Huang_2017} & 91.93 & 92.20 & 91.93 & 91.89 & 88.67 & 89.23 & 88.67 & 88.67 & 86.81 & 88.24 & 86.81 & 86.71 \\
ViT \citep{Dosovitskiy_2020} & 84.59 & 85.59 & 84.59 & 84.67 & 74.81 & 77.53 & 74.81 & 75.01 & 65.19 & 65.72 & 65.19 & 63.92 \\
Swin Transformer \citep{Liu_2021} & 92.67 & 92.69 & 92.67 & 92.63 & 91.41 & 91.42 & 91.41 & 91.36 & 85.56 & 85.80 & 85.56 & 85.27 \\
Mean Teacher \citep{Tarvainen_2017} & 93.74 & 93.84 & 93.74 & 93.71 & 91.80 & 91.82 & 91.80 & 91.65 & 83.63 & 83.96 & 83.63 & 83.55 \\
FixMatch \citep{Sohn_2020_FixMatch} & 94.63 & 91.88 & 92.30 & 91.92 & 91.95 & 88.66 & 89.94 & 89.04 & 89.12 & 86.75 & 85.16 & 85.65 \\
FlexMatch \citep{Zhang_2021_FlexMatch} & 91.78 & 91.04 & 90.64 & 91.00 & 89.87 & 86.98 & 86.73 & 86.05 & 86.29 & 82.60 & 81.42 & 80.94 \\
SoftMatch \citep{Chen_2023_SoftMatch} & 94.04 & 94.20 & 94.04 & 93.99 & 91.51 & 91.99 & 91.51 & 91.55 & 89.12 & 89.31 & 89.12 & 88.98 \\
GapMatch \citep{Huang_2025} & 94.90 & 93.82 & 94.90 & 93.17 & 92.04 & 92.54 & 92.01 & 91.78 & 90.26 & 89.49 & 90.26 & 89.21 \\

\midrule
\textbf{U-SSDNet (Proposed)} & \textbf{95.07} & \textbf{94.73} & \textbf{94.95} & \textbf{94.73}
& \textbf{92.85} & \textbf{92.98} & \textbf{92.25} & \textbf{92.95}
& \textbf{90.87} & \textbf{90.16} & \textbf{90.42} & \textbf{89.88} \\
\bottomrule
\end{tabular}%
}
\end{table}

\subsubsection{Abalation Study}
\begin{figure*}[htbp]
    \centering
    \subfigure[Augmentation module]{
        \includegraphics[width=0.23\textwidth, height=3cm]{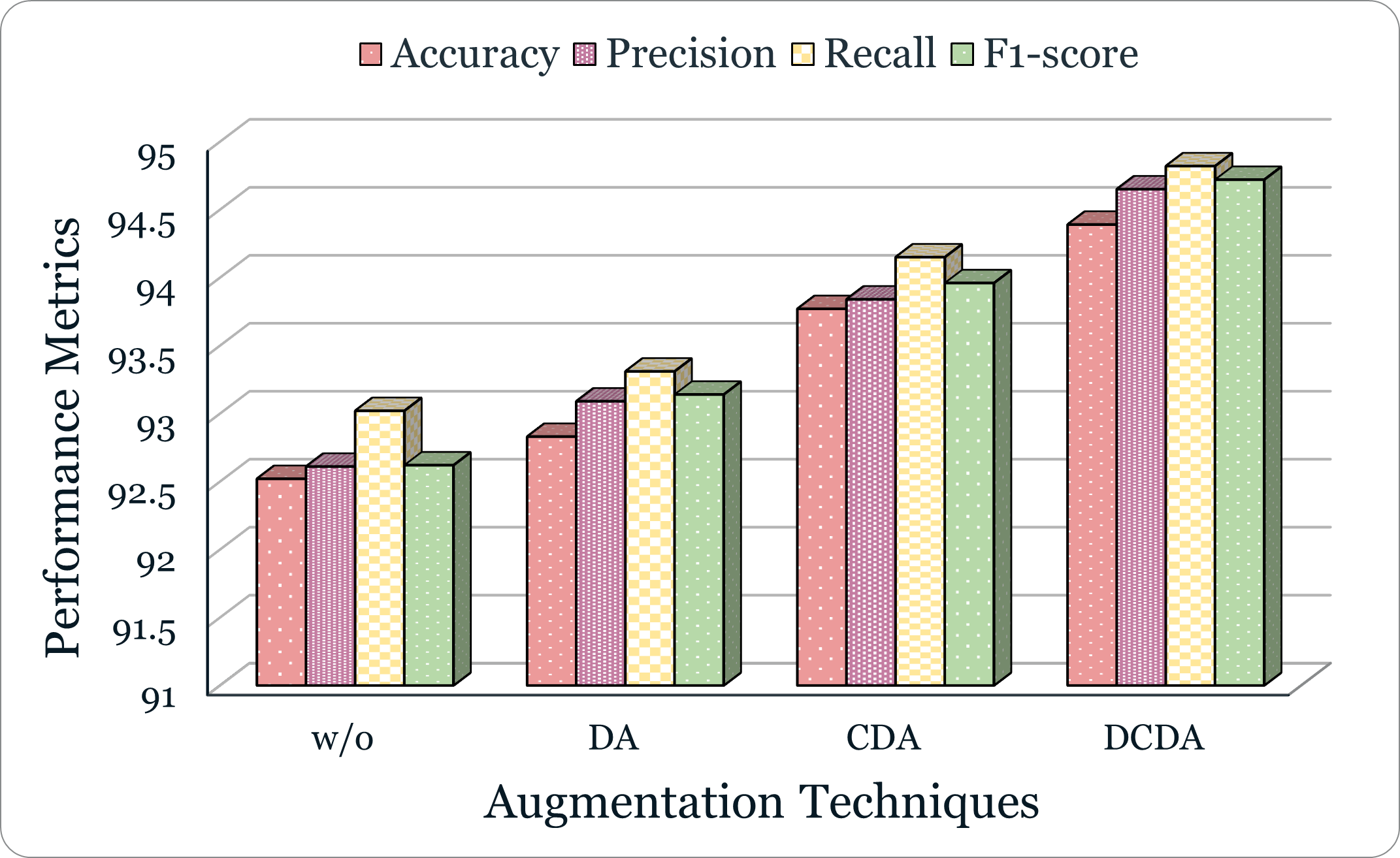}
        \label{fig:Aug_SLD-1}
    }
    \hfill
    \subfigure[Feature Encoding module]{
        \includegraphics[width=0.23\textwidth, height=3cm]{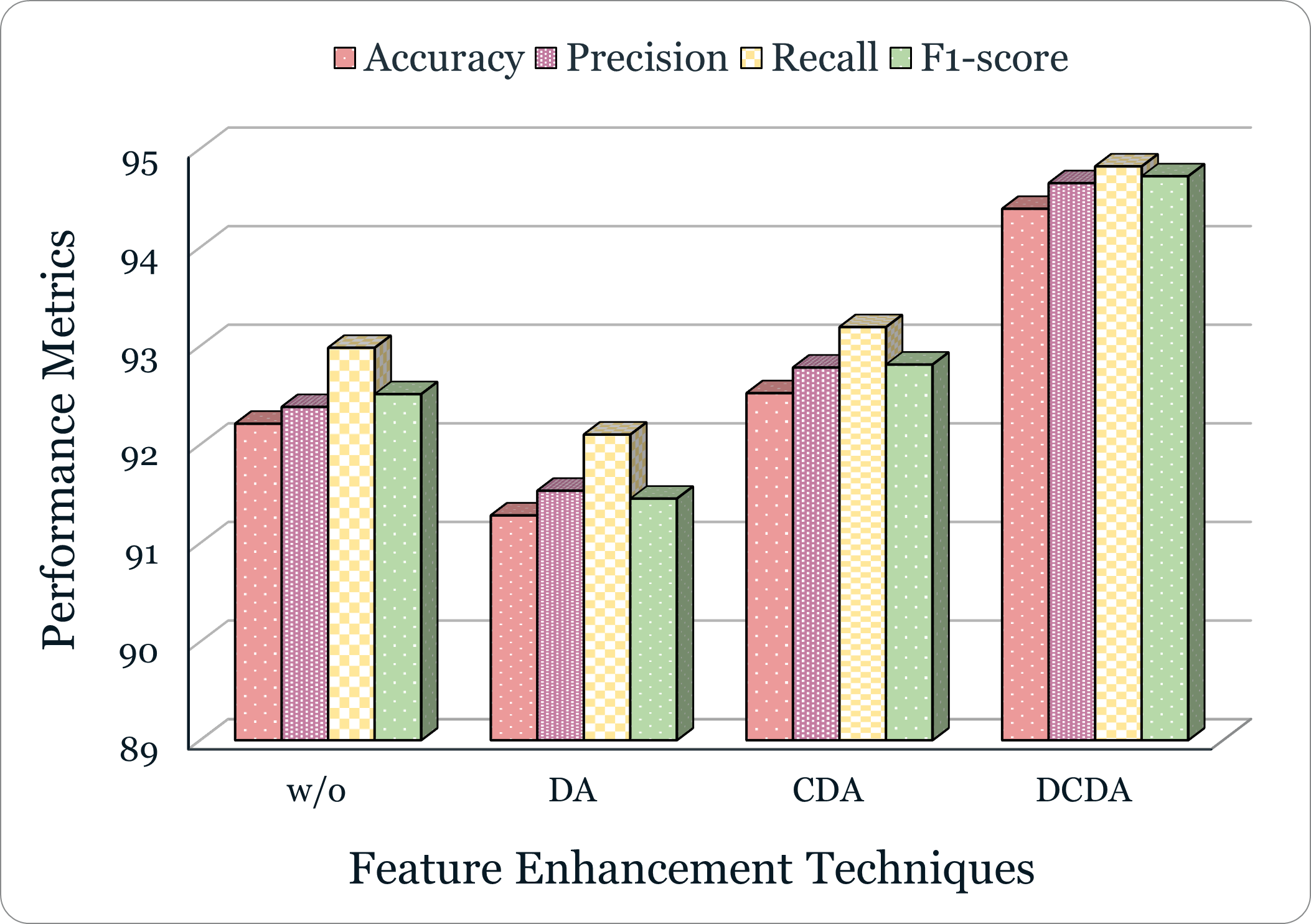}
        \label{fig:FEn_SLD-1}
    }
    \hfill
    \subfigure[Feature Extraction module]{
        \includegraphics[width=0.23\textwidth, height=3cm]{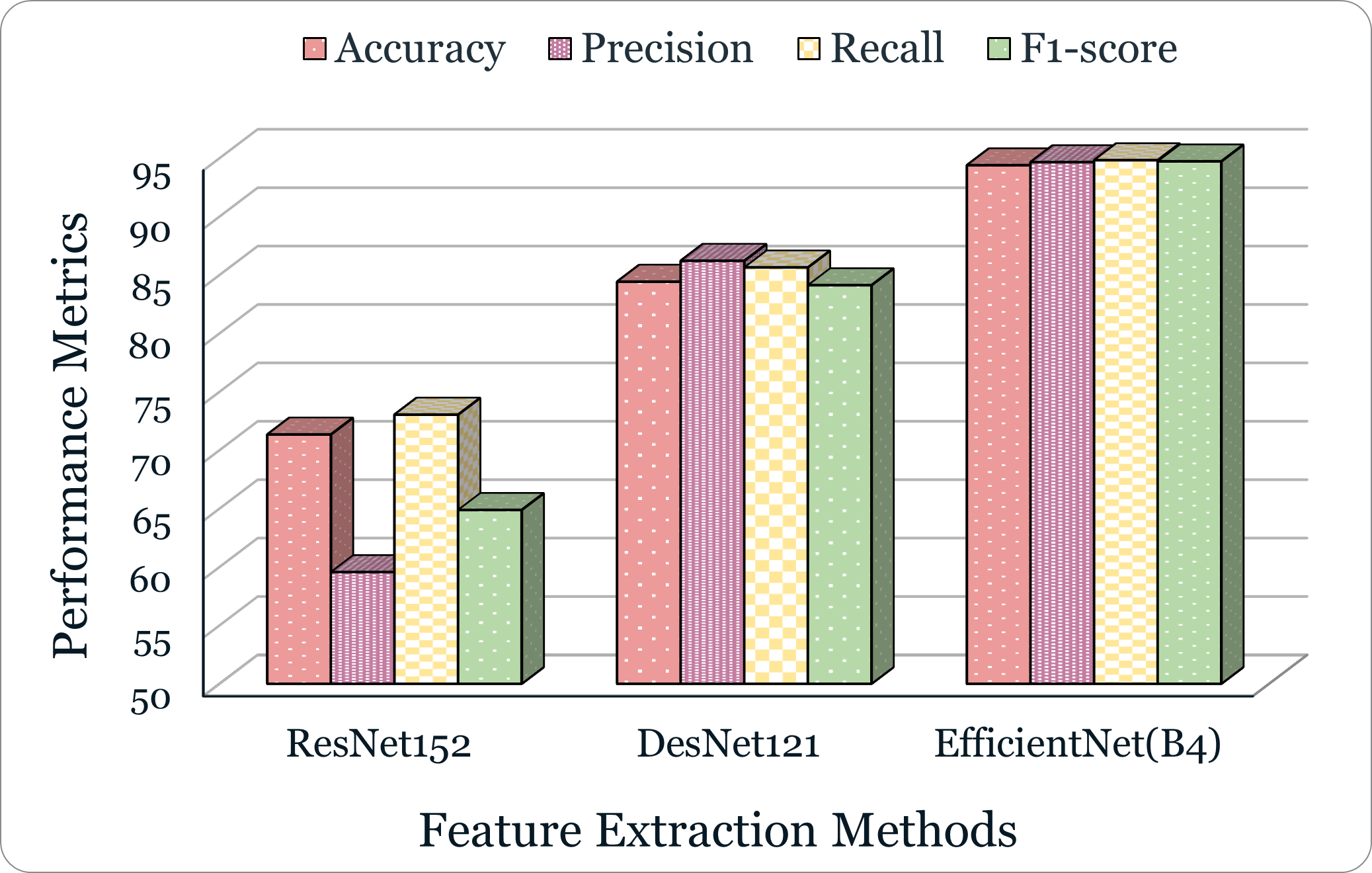}
        \label{fig:FEx_SLD-1}
    }
    \hfill
    \subfigure[Uncertainty Aware module]{
        \includegraphics[width=0.23\textwidth, height=3cm]{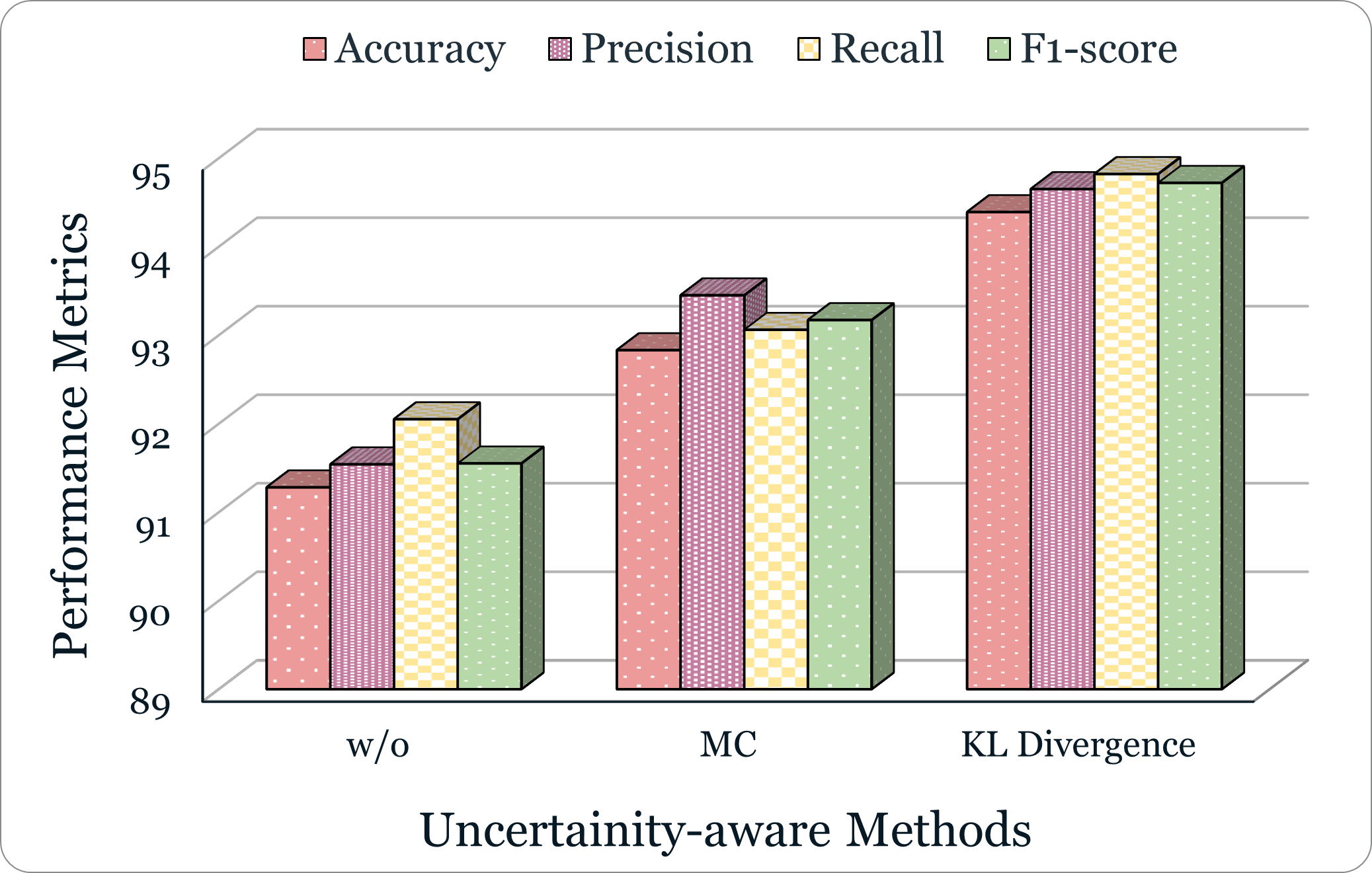}
        \label{fig:UA_SLD-1}
    }
    \caption{Ablation study results on the Sugarcane Leaf Disease (SLD-1) dataset on various components such as: (a) Augmentation module; (b) Feature Encoding module; (c) Feature Extraction module; and (d) Uncertainty-Aware module.}
    \label{ablations}
\end{figure*}

We performed an ablation study in which we removed individual modules and analyzed their impact on the sugarcane leaf disease dataset to evaluate the contributions of each key module in our proposed framework, as shown in~\autoref{ablations}. Our findings indicate that each component and strategy significantly improves the performance of our method. We precisely analyzed the functions of Leaf Image Preprocessing and Augmentation, the Semi-Supervised Mean-Teacher Model, and Uncertainty Estimation. Section \ref{Methodology} gives a detailed overview of the proposed model pipeline.
The ablation study evaluates the impact of different feature extraction techniques on the performance of the proposed U-SSDNet model, as defined in~\autoref{tab:fe}.
 The analysis shows that EfficientNet consistently achieves outstanding results across all evaluation metrics and selects it as the most effective architecture among those examined. 
DenseNet performs better than ResNet, but still lags behind EfficientNet. So the backbone choice is very important for optimizing U-SSDNet, and EfficientNet provides the best feature representation.
\paragraph{Impact of Feature Enhancement}
~\autoref{tab:wavelet} shows that different feature enhancement techniques affect the performance of the U-SSDNet model. Among these techniques, the Wavelet-based enhancement achieves the best results with the metrics accuracy, precision, recall, and F1-score. The HOG and LPB methods show lower performance, with accuracy gaps of 3.52\% and 2.28\%, respectively. These results show that Wavelet-based feature enhancement enables the model to capture important structural and textural details of the leaf and to improve classification.

\begin{table}[htbp]\small
\centering
\caption{Comparative analysis of U-SSDNet model performance across different feature enhancement methods. The ``diff'' columns represent the performance gap relative to the best-performing configuration (Wavelet).}
\label{tab:wavelet}
\begin{tabular*}{\linewidth}{@{\extracolsep{\fill}} l c c c c c c c c}
\toprule
\textbf{Model} & \textbf{Acc} & \textbf{diff} & \textbf{Pre} & \textbf{diff} & \textbf{Rec} & \textbf{diff} & \textbf{F1} & \textbf{diff} \\
\midrule
U-SSDNet (w/o) & 92.21 & 2.59 & 92.38 & 2.44 & 92.98 & 1.90 & 92.51 & 1.69 \\
U-SSDNet (HOG) & 91.28 & 3.52 & 91.53 & 3.29 & 92.11 & 2.77 & 91.45 & 2.75 \\
U-SSDNet (LPB) & 92.52 & 2.28 & 92.78 & 2.04 & 93.19 & 1.69 & 92.81 & 1.39 \\
\midrule
\textbf{U-SSDNet (Wavelet)} & \textbf{94.80} & \textbf{0.00} & \textbf{94.82} & \textbf{0.00} & \textbf{94.88} & \textbf{0.00} & \textbf{94.20} & \textbf{0.00} \\
\bottomrule
\end{tabular*}
\end{table}

\paragraph{Impact of Augmentation Techniques}
~\autoref{tab:augmentation_comparison} shows that combining Deep and Cut\&Delete augmentations gives the best U-SSDNet performance, achieving the highest accuracy (94.80\%), precision (94.82\%), recall (94.88\%), and F1-score (94.20\%). Augmentations perform slightly better, while the model without augmentation shows the weakest results, confirming that combining augmentation techniques enhances model generalization and classification accuracy.

\begin{table}[htbp]\small
\centering
\caption{Comparative analysis of U-SSDNet model performance across different augmentation techniques. The ``diff'' columns indicate the performance gap relative to the best-performing configuration (Deep+Cut\&Delete).}
\label{tab:augmentation_comparison}
\begin{tabular*}{\linewidth}{@{\extracolsep{\fill}} l c c c c c c c c}
\toprule
\textbf{Model} & \textbf{Acc} & \textbf{diff} & \textbf{Pre} & \textbf{diff} & \textbf{Rec} & \textbf{diff} & \textbf{F1} & \textbf{diff} \\
\midrule
U-SSDNet (w/o Aug) & 92.52 & 2.28 & 92.61 & 2.21 & 93.02 & 1.86 & 92.62 & 1.58 \\
U-SSDNet (Deep Aug) & 92.83 & 1.97 & 93.09 & 1.73 & 93.31 & 1.57 & 93.14 & 1.06 \\
U-SSDNet (Cut\&Delete Aug) & 93.77 & 1.03 & 93.84 & 0.98 & 94.15 & 0.73 & 93.96 & 0.24 \\
\midrule
\textbf{U-SSDNet (Deep+Cut\&Delete)} & \textbf{94.80} & \textbf{0.00} & \textbf{94.82} & \textbf{0.00} & \textbf{94.88} & \textbf{0.00} & \textbf{94.20} & \textbf{0.00} \\
\bottomrule
\end{tabular*}
\end{table}

\paragraph{Impact of Feature Extraction Approaches}
~\autoref{tab:fe} shows that the choice of feature extraction method greatly affects the performance of the U-SSDNet model. The EfficientNet backbone achieves the best performance with an accuracy of 94.80\%, precision of 94.82\%, recall of 94.88\%, and an F1-score of 94.20\%. while ResNet and DenseNet show considerably lower results with accuracies of 71.34\% \% and 84.42\%. The results show that EfficientNet captures more relevant and discriminative features, leading to improved classification performance.

\begin{table}[htbp]\small
\centering
\caption{Comparative analysis of U-SSDNet model performance across different feature extraction methods. The ``diff'' columns represent the performance gap relative to the best-performing configuration (EfficientNet).}
\label{tab:fe}
\begin{tabular*}{\linewidth}{@{\extracolsep{\fill}} l c c c c c c c c}
\toprule
\textbf{Model} & \textbf{Acc} & \textbf{diff} & \textbf{Pre} & \textbf{diff} & \textbf{Rec} & \textbf{diff} & \textbf{F1} & \textbf{diff} \\
\midrule
U-SSDNet (ResNet) & 71.34 & 23.46 & 59.56 & 35.26 & 73.04 & 21.84 & 64.88 & 29.32 \\
U-SSDNet (DenseNet) & 84.42 & 10.38 & 86.21 & 8.61 & 85.64 & 9.24 & 84.12 & 10.08 \\
\midrule
\textbf{U-SSDNet (EfficientNet)} & \textbf{94.80} & \textbf{0.00} & \textbf{94.82} & \textbf{0.00} & \textbf{94.88} & \textbf{0.00} & \textbf{94.20} & \textbf{0.00} \\
\bottomrule
\end{tabular*}
\end{table}

\paragraph{Impact of Uncertainty-aware Methods}
\autoref{tab:uncertainty} shows that using an uncertainty-aware approach improves the performance of the U-SSDNet model. The models without uncertainty awareness (w/o UA) and the MC-based UA approach achieved lower accuracy than the KL-Divergence method. This result shows that the KL-Divergence method effectively handles prediction uncertainty, improving the model's stability and precision.

\begin{table}[htbp]\small
\centering
\caption{Comparative analysis of U-SSDNet performance across different uncertainty-aware methods. The ``diff'' columns represent the performance gap relative to the best-performing configuration (KL-Divergence).}
\label{tab:uncertainty}
\begin{tabular*}{\linewidth}{@{\extracolsep{\fill}} l c c c c c c c c}
\toprule
\textbf{Model} & \textbf{Acc} & \textbf{diff} & \textbf{Pre} & \textbf{diff} & \textbf{Rec} & \textbf{diff} & \textbf{F1} & \textbf{diff} \\
\midrule
U-SSDNet (w/o UA) & 91.28 & 3.52 & 91.54 & 3.28 & 92.05 & 2.83 & 91.55 & 2.65 \\
U-SSDNet (MC) & 92.83 & 1.97 & 93.45 & 1.37 & 93.06 & 1.82 & 93.17 & 1.03 \\
\midrule
\textbf{U-SSDNet (KL-Divergence)} & \textbf{94.80} & \textbf{0.00} & \textbf{94.82} & \textbf{0.00} & \textbf{94.88} & \textbf{0.00} & \textbf{94.20} & \textbf{0.00} \\
\bottomrule
\end{tabular*}
\end{table}

\begin{figure*}[h!]
    \centering
    \includegraphics[width=16cm, height=10cm]{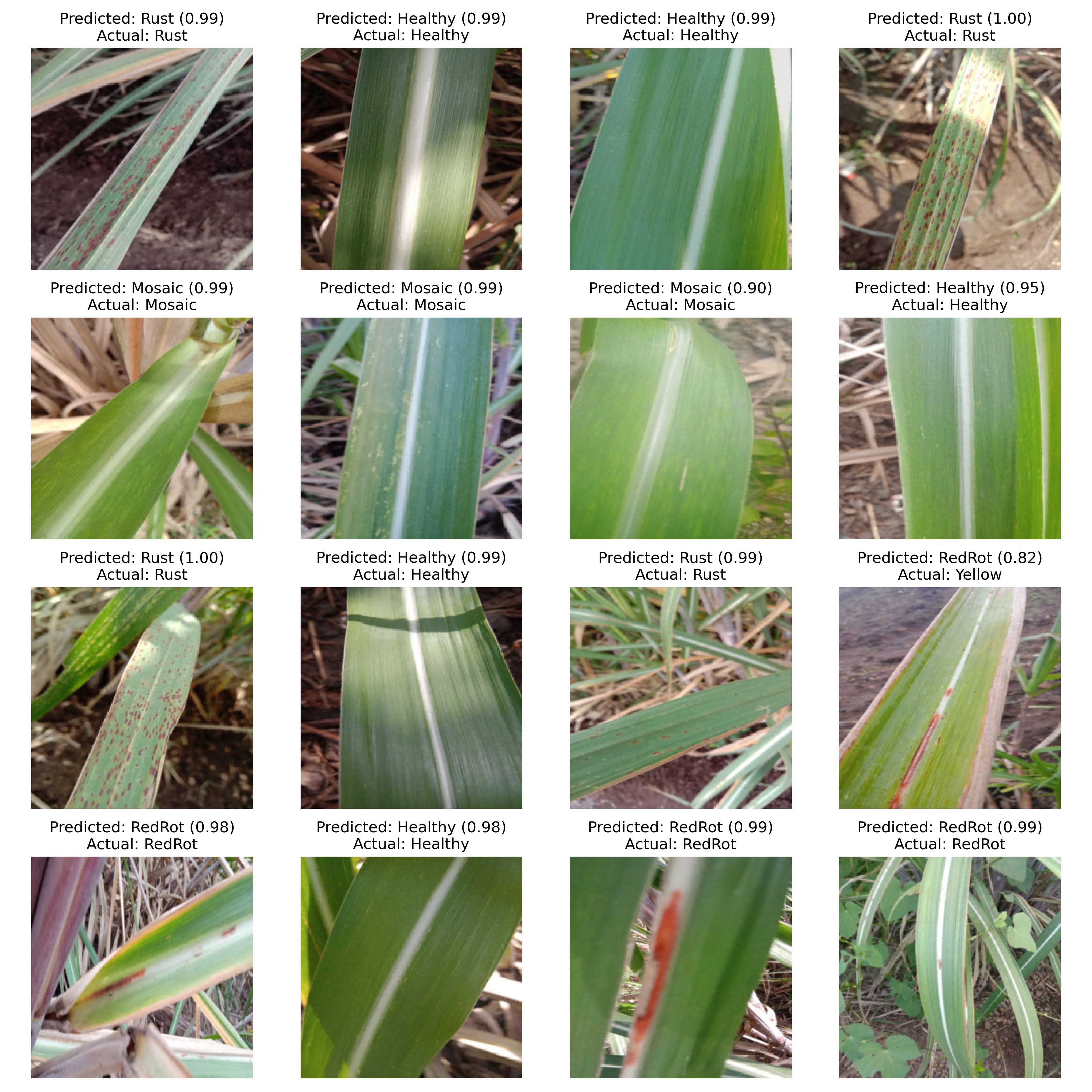}
    \caption{Prediction results of sugarcane leaf diseases on the SLD-1 dataset.}
    \label{QA_SLD1}
\end{figure*}
\subsubsection{Qualitative Analysis}
\begin{figure*}
    \centering
    \includegraphics[width=16cm, height=10cm]{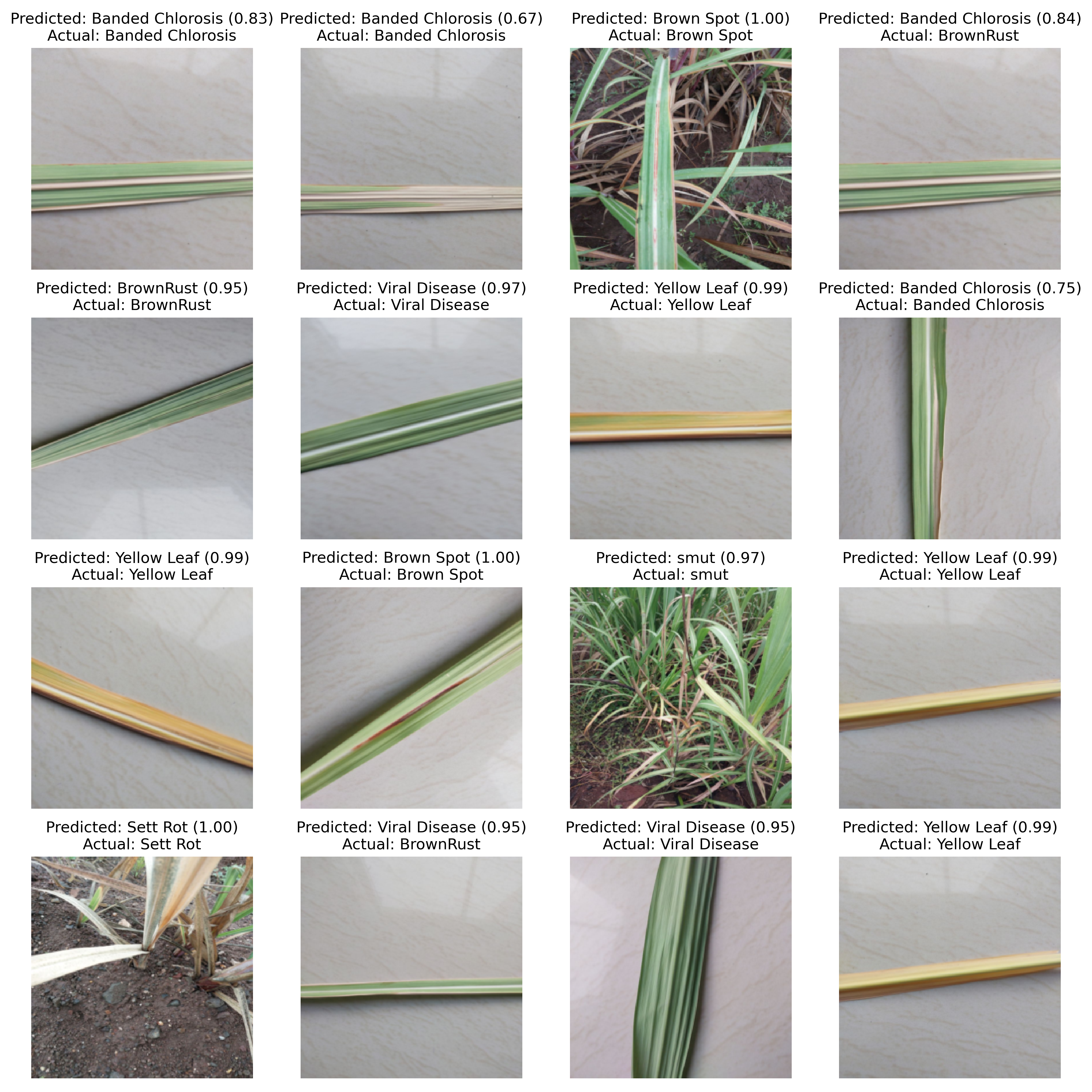}
    \caption{Prediction results of sugarcane leaf diseases on the SLD-2 dataset.}
    \label{QA_SLD2}
\end{figure*}

The qualitative evaluation of the U-SSDNet method on the SLD-1 and SLD-2 datasets is shown in \autoref{QA_SLD1} and \autoref{QA_SLD2}, which focus on its prediction behavior across different leaf disease classes. The model shows consistently high confidence for most samples and correctly classifies healthy, Redrot, Mosaic, and Rust leaves with good prediction scores. Most samples achieve prediction confidence above 0.9 and indicate the model's reliability in distinguishing between different disease categories. Only a few misclassifications are observed, such as one Yellow leaf sample is incorrectly predicted as Redrot with a confidence of 0.82. But these cases are rare and generated because the same feature is contained in two different leaf disease classes. These results show that U-SSDNet not only achieves high performance but also produces convincing predictions and maintains its effectiveness for leaf disease classification in SLD-1.

U-SSDNet shows better performance in precisely classifying sugarcane leaf diseases over a wide range of samples. Most of the predictions have confidence scores above 0.95, which shows strong reliability. Leaf diseases such as Brown Spot, Sett Rot, Viral Disease, and Yellow Leaf are accurately predicted and match the actual results. Even for similar diseases such as Banded Chlorosis and Brown Rust, the model maintains confidence scores above 0.75, and the few errors observed have little effect on overall performance. The findings highlight the proposed model's ability to capture different features across patterns in leaf texture, color, and appearance. Consistently confident predictions indicate the model's reliability.

\begin{figure*}[htbp]
    \centering
    \includegraphics[width=15cm, height=9.85cm]{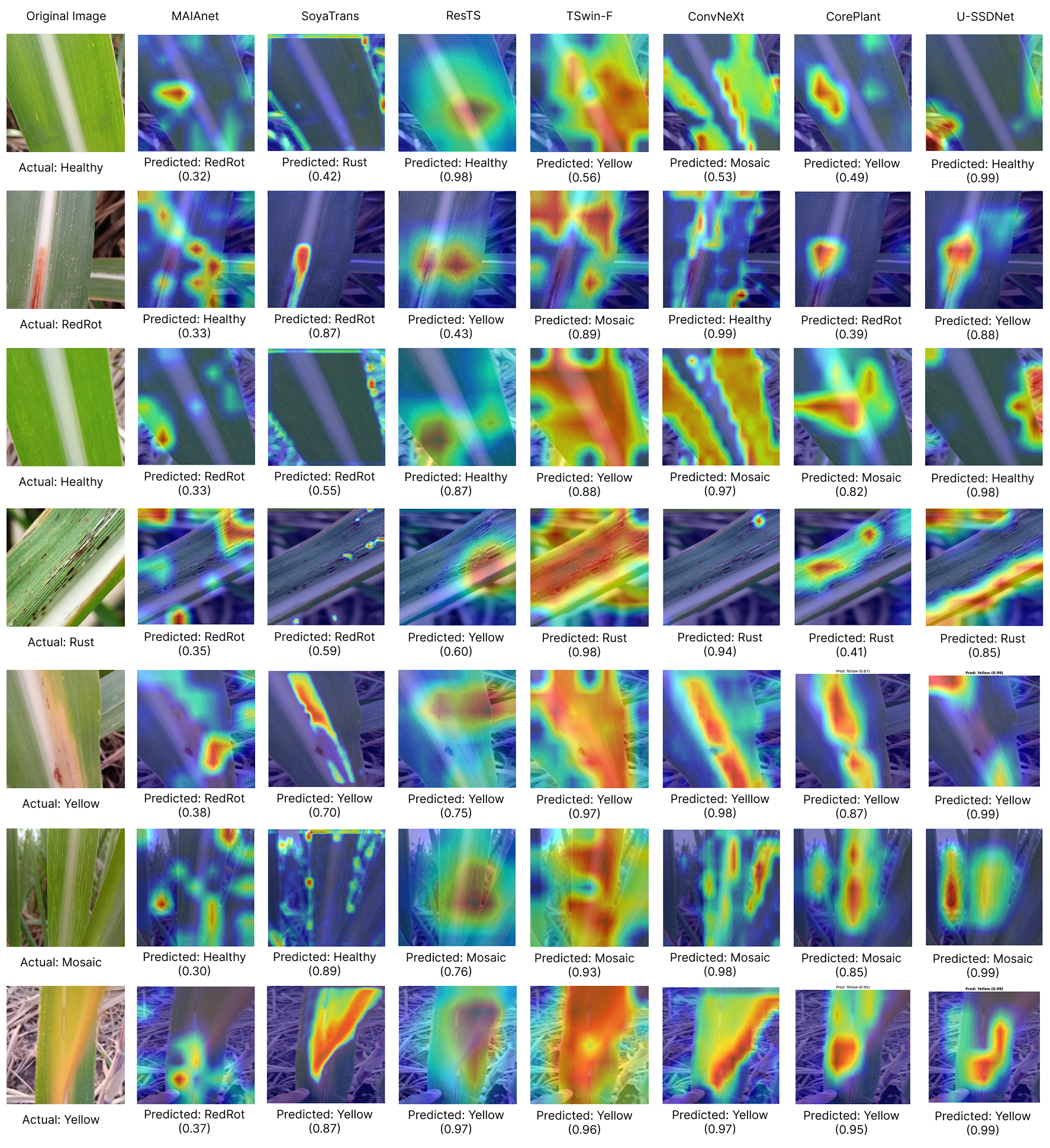}
    \caption{Comparison of Grad-CAM visualizations for different methods on the SLD-1 dataset (10\% labeled data).}
    \label{GD-1}
\end{figure*}

\begin{figure*}[htbp]
    \centering
    \includegraphics[width=15cm, height=9.85cm]{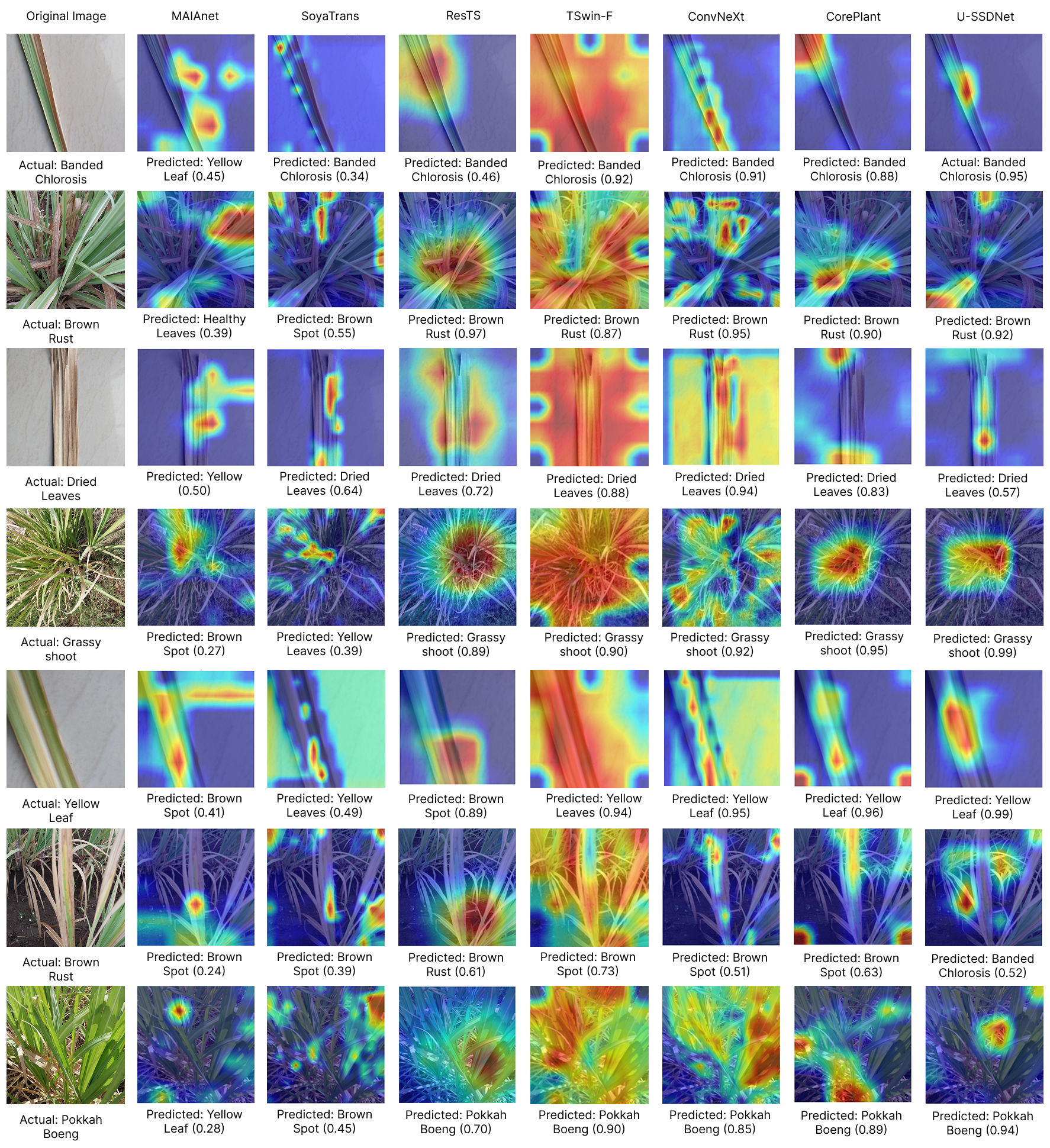}
    \caption{Comparison of Grad-CAM visualizations for different methods on the SLD-2 dataset (10\% labeled data).}
    \label{GD-2}
\end{figure*}

\begin{figure*}[htbp]
    \centering
    \includegraphics[width=15cm, height=9.85cm]{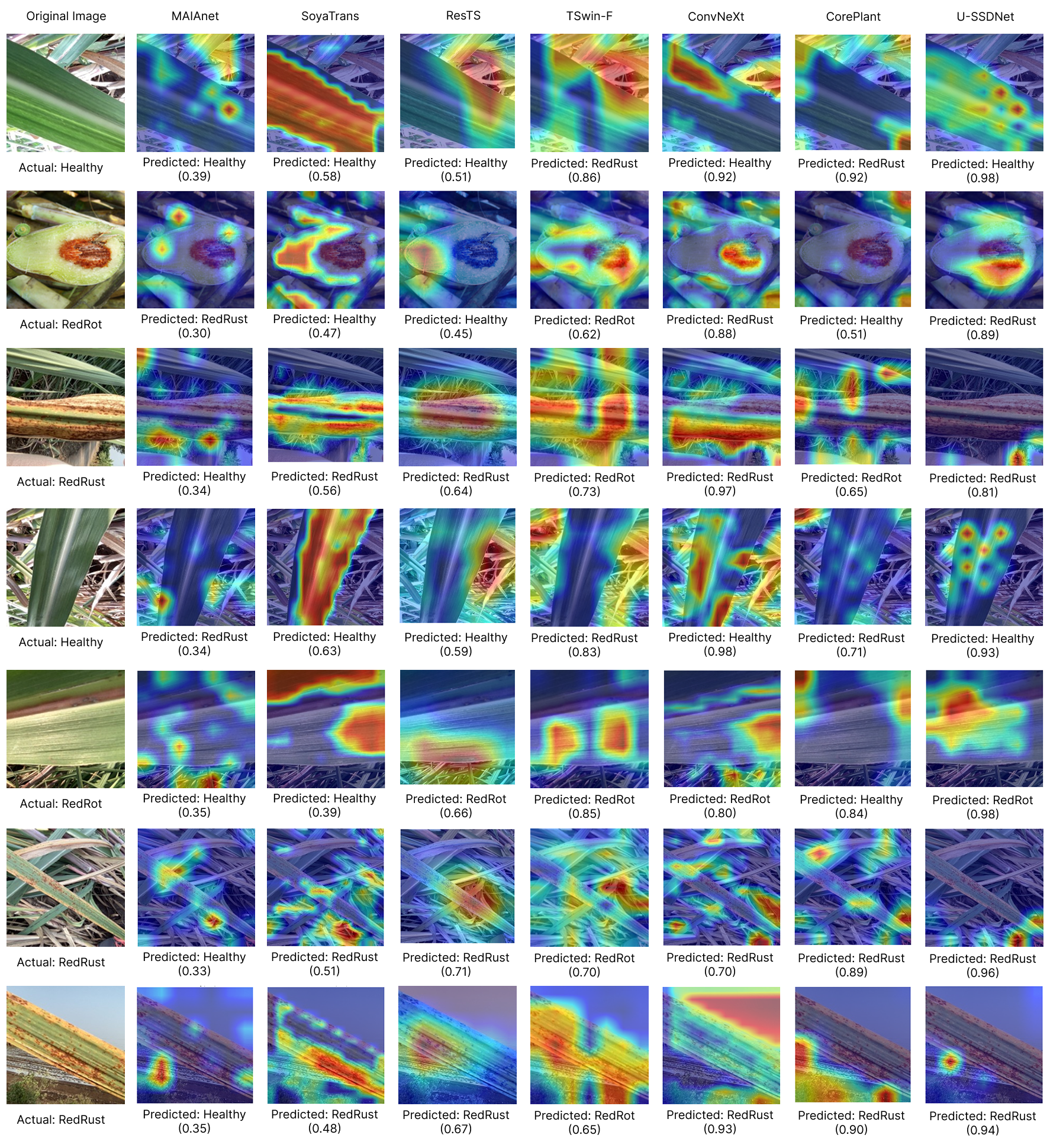}
    \caption{Comparison of Grad-CAM visualizations for different methods on the SLD-3 dataset (10\% labeled data).}
    \label{GD-3}
\end{figure*}

\subsubsection{Qualitative Comparison with Existing Methods}
\autoref{GD-1}--\autoref{GD-3} present Grad-CAM visualizations for qualitative comparison of different state-of-the-art methods on the SLD-1, SLD-2, and SLD-3 datasets using 10\% labeled training data. The visualizations emphasize the regions contributing most to model predictions and provide insights into the discriminatory features learned under limited supervision. Compared with U-SSDNet, baseline models exhibit distributed attention patterns that do not consistently align with disease symptoms.

Several competing models struggle to distinguish visually similar disease categories, leading to inconsistent localization of disease symptoms in the SLD-1 dataset, as shown in~\autoref{GD-1}. On the SLD-2 dataset, attention maps from baseline approaches frequently extend beyond symptomatic regions, indicating reliance on less informative features, as shown in~\autoref{GD-2}. In contrast, U-SSDNet consistently focuses on disease-relevant regions, producing more localized, semantically meaningful activations that closely align with visible symptoms.

A comparable pattern is observed on the SLD-3 dataset, where subtle inter-class differences make disease recognition challenging, as shown in~\autoref{GD-3}. While baseline models often generate fragmented attention patterns, U-SSDNet effectively highlights attention on characteristic disease regions, leading to more reliable predictions. The Grad-CAM visualizations show that U-SSDNet focuses more consistently on disease-specific regions than competing methods.

\subsubsection{Class-wise Performance Assessment}
\begin{figure*}[ht]
    \centering
    \includegraphics[width=15cm, height=10cm]{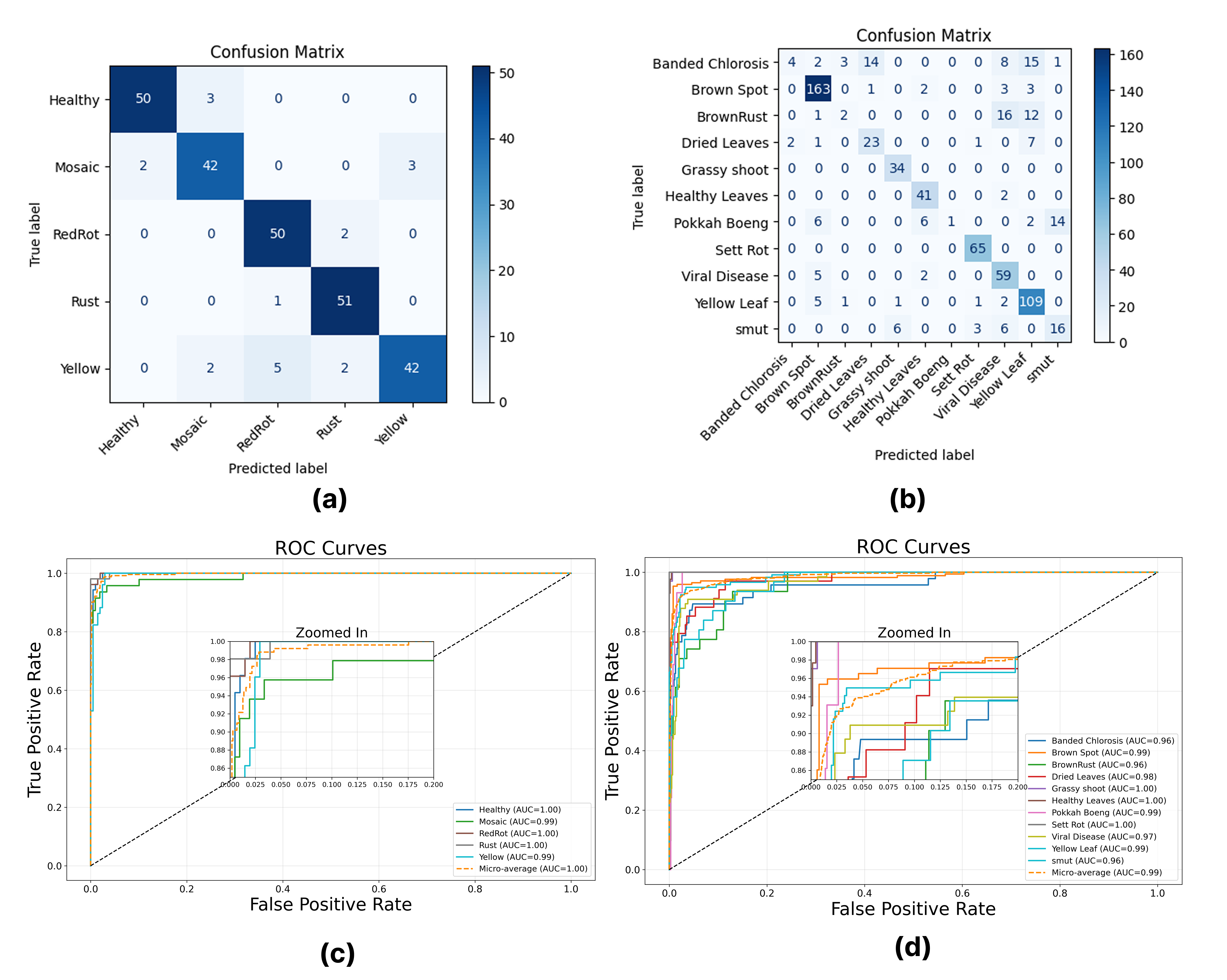}
    \caption{Performance across various classes in the SLD-1 and SLD-2 datasets. Figures (a) and (b) show the confusion matrices of U-SSDNet for SLD-1 and SLD-2, respectively, while Figures (c) and (d) present the corresponding ROC curves for both datasets.}

    \label{roc_cn}
\end{figure*}

{We perform a class-wise evaluation of the model on the SLD-1 and SLD-2 datasets, as shown in \autoref{roc_cn}. This includes confusion matrices and Receiver Operating Characteristic (ROC) curves for each class, with the Area Under the Curve (AUC) used as the main metric. \autoref{roc_cn}(a) and (b) show the confusion matrices for SLD-1 and SLD-2, indicating how well the model correctly classifies each disease and where errors occur. \autoref{roc_cn}(c) and (d) present the corresponding ROC curves, showing the model's ability to distinguish between classes. High AUC values indicate strong performance, while lower values highlight more challenging classes with similar leaf disease patterns. This class-wise analysis provides a clear view of the model's performance across all disease categories.}

\section{Discussion}
\label{Discussion}

\subsection{CAM Visualization}
\begin{figure}[h!]
    \centering
    \includegraphics[width=15cm, height=8cm]{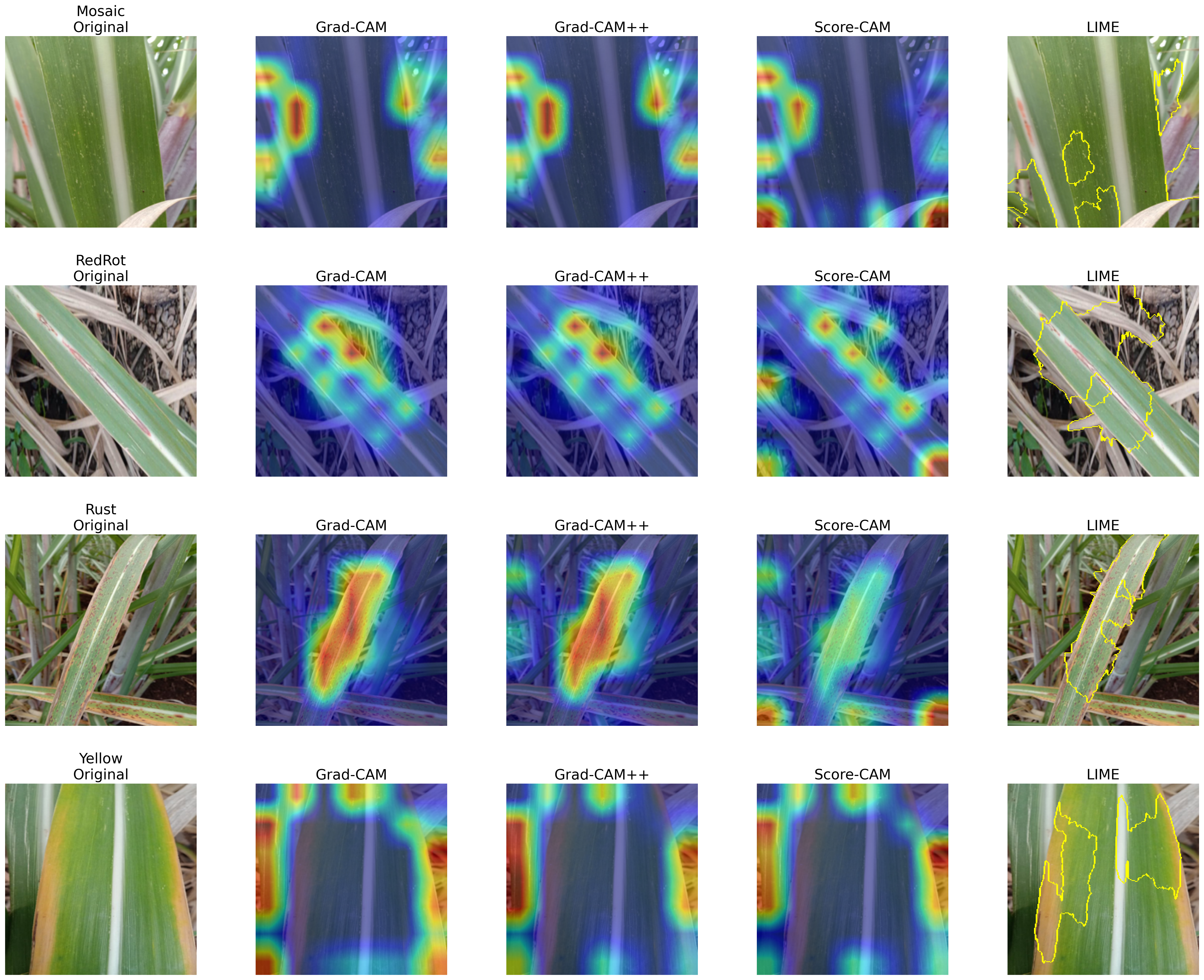}
    \caption{This figure presents a comparison of CAM and LIME visualizations on the SLD-1 dataset for different diseases and shows the effectiveness of the proposed method in identifying disease-affected regions.}
    \label{fig:CAM_comparison}
\end{figure}

\autoref{fig:CAM_comparison} displays activation maps produced by four well-known visualization methods, Grad-CAM, Grad-CAM++, Score-CAM, and LIME, to expose the regions most influential in the mean-teacher model's decisions. For each disease category, Mosaic, RedRot, Rust, and Yellow, the first column shows a representative test image, while the remaining columns display the corresponding activation maps. These techniques show which parts of the image most affect the model's predictions. Grad-CAM provides a broad overview of discriminative regions, while Grad-CAM++ produces sharper, more precisely localized regions. The visualizations demonstrate that the proposed model consistently focuses on the true disease-affected regions of sugarcane leaves, even when symptoms are subtle or low in intensity. The highlighted regions coincide well with visible lesion patterns and show that the network can detect early signs of infection and recognize fine structural details in Mosaic and RedRot samples. Grad-CAM++ in particular shows clear boundaries around the symptomatic areas, confirming that the learned features are biologically meaningful and not merely correlated background artifacts.
Some variability is observed in Rust leaves, where the model occasionally concentrates on only part of the infected area. This reflects the greater difficulty in detecting scattered, small pustules compared to more extensive lesions. However, the overall consistency across the three visualization methods provides strong evidence that the mean-teacher classifier reliably identifies discriminative patterns. 

\subsection{t-SNE Visualization}
\begin{figure}[h!]
    \centering
    \subfigure[Before training]{
        \includegraphics[width=0.31\textwidth, height=3.5cm]{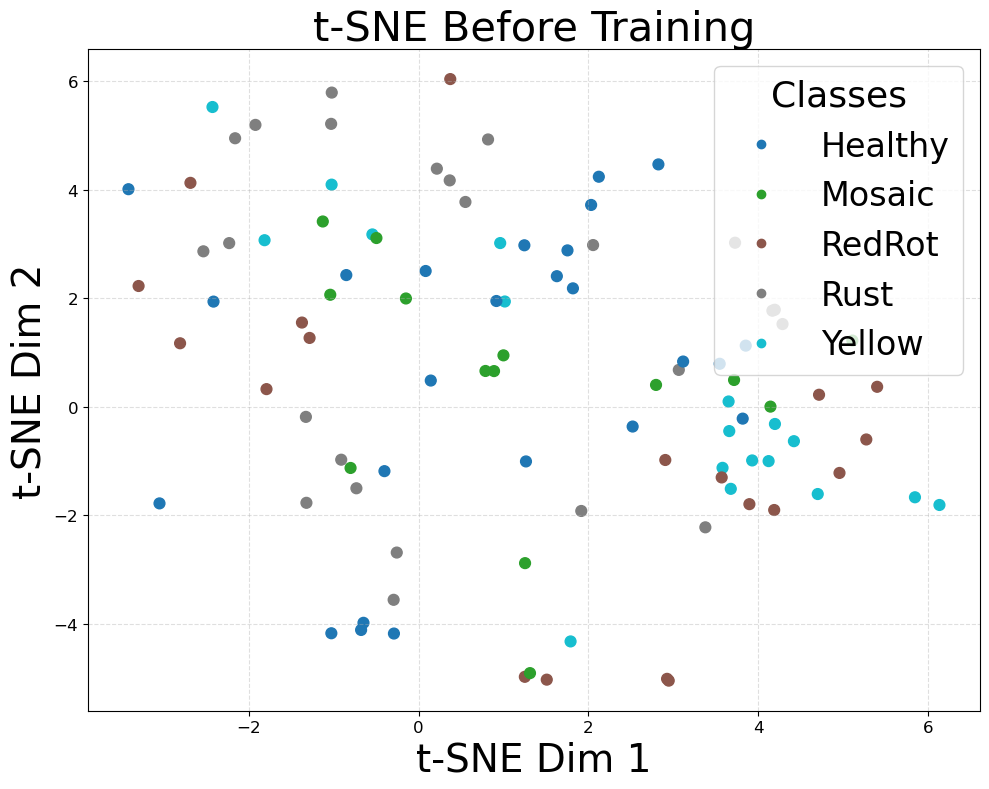}
        \label{fig:roc_SLD-1}}
    \hfill
    \subfigure[10\% labeled]{
        \includegraphics[width=0.31\textwidth, height=3.5cm]{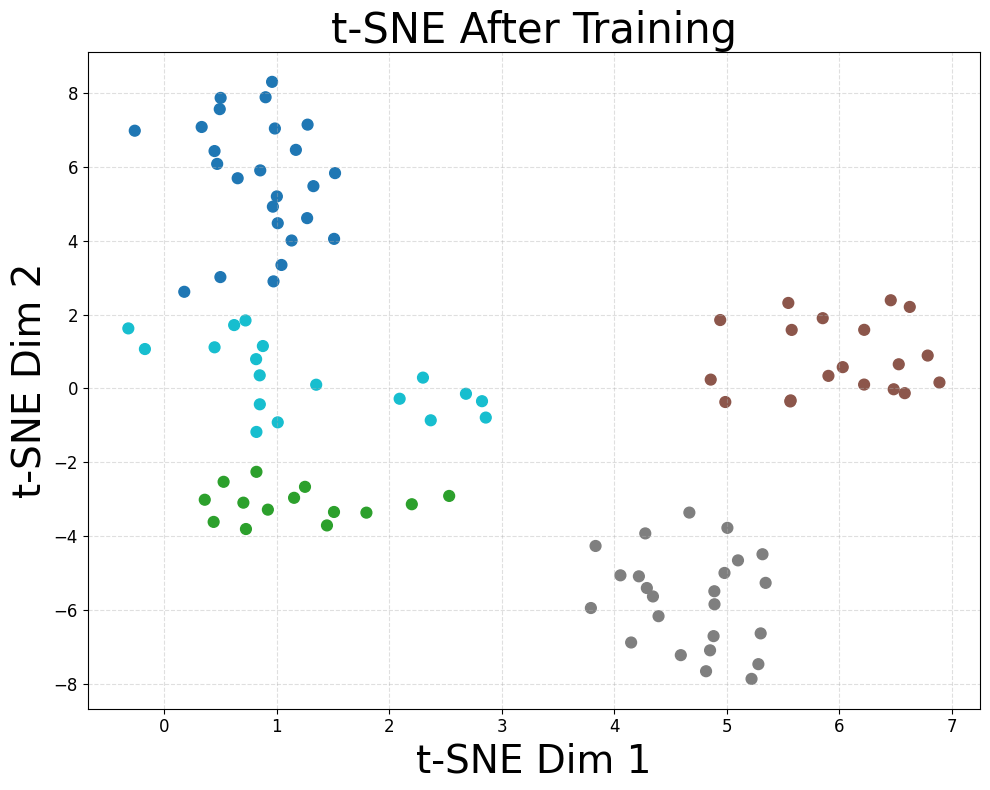}
        \label{fig:Pred_SLD-1_10}}
    \hfill
    \subfigure[20\% labeled]{
        \includegraphics[width=0.31\textwidth, height=3.5cm]{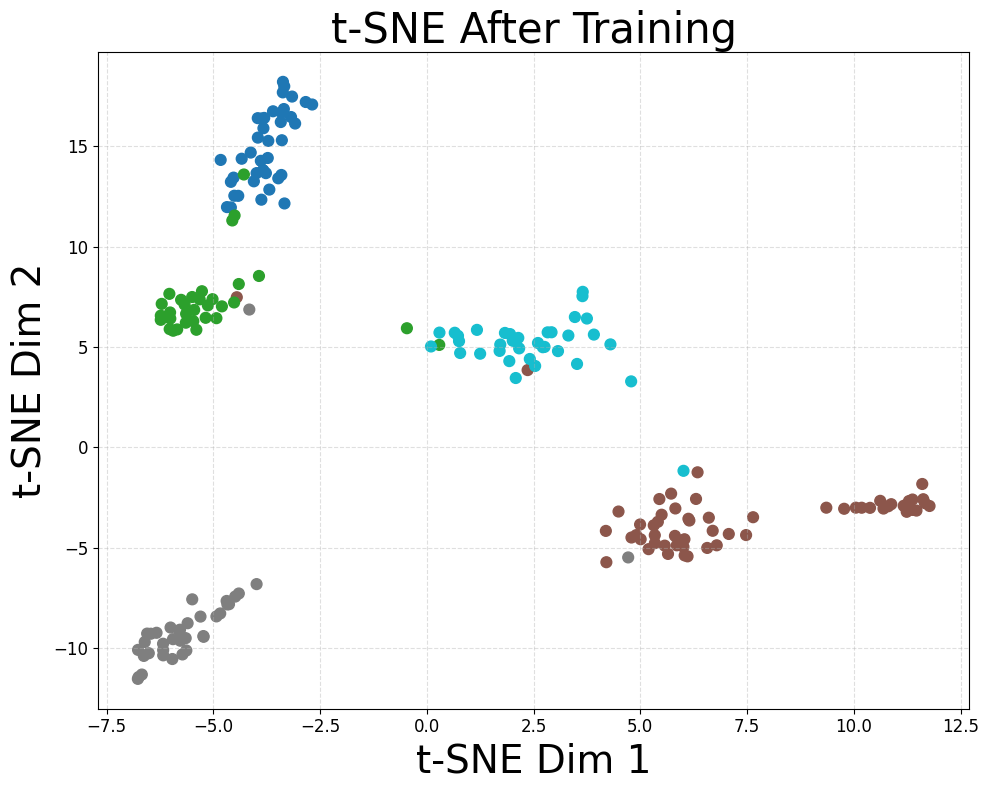}
        \label{fig:Pred_SLD-1_20}}
    \hfill
    \subfigure[30\% labeled]{
        \includegraphics[width=0.31\textwidth, height=3.5cm]{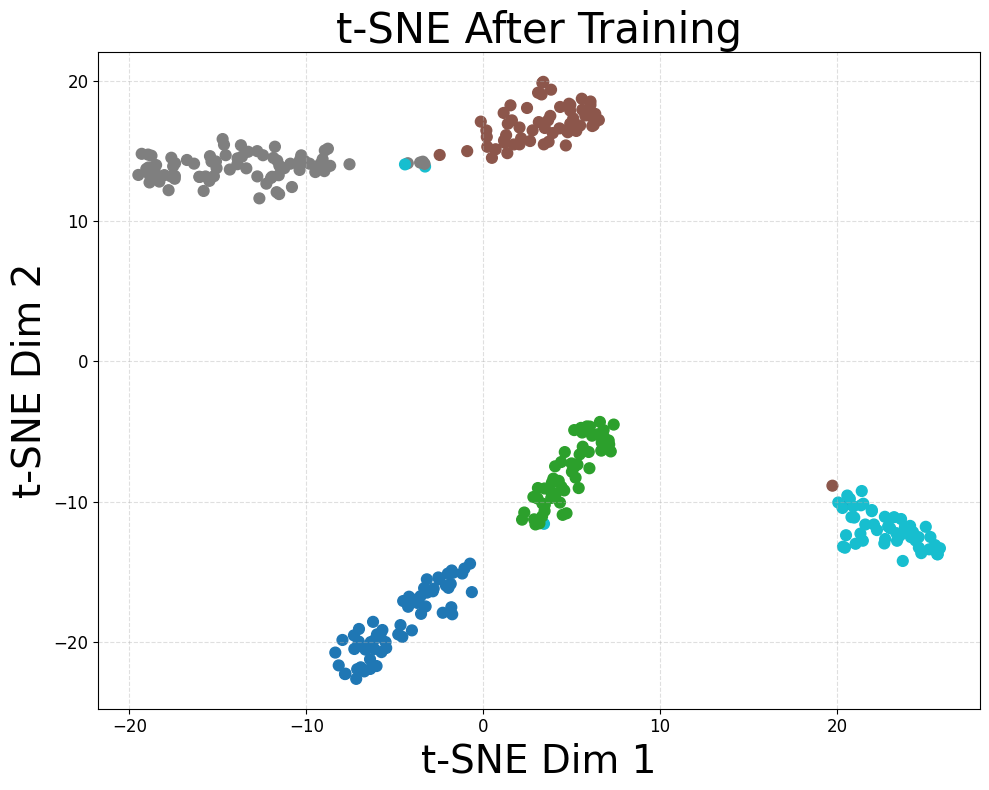}
        \label{fig:Pred_SLD-1_30}}
    \hfill
    \subfigure[40\% labeled]{
        \includegraphics[width=0.31\textwidth, height=3.5cm]{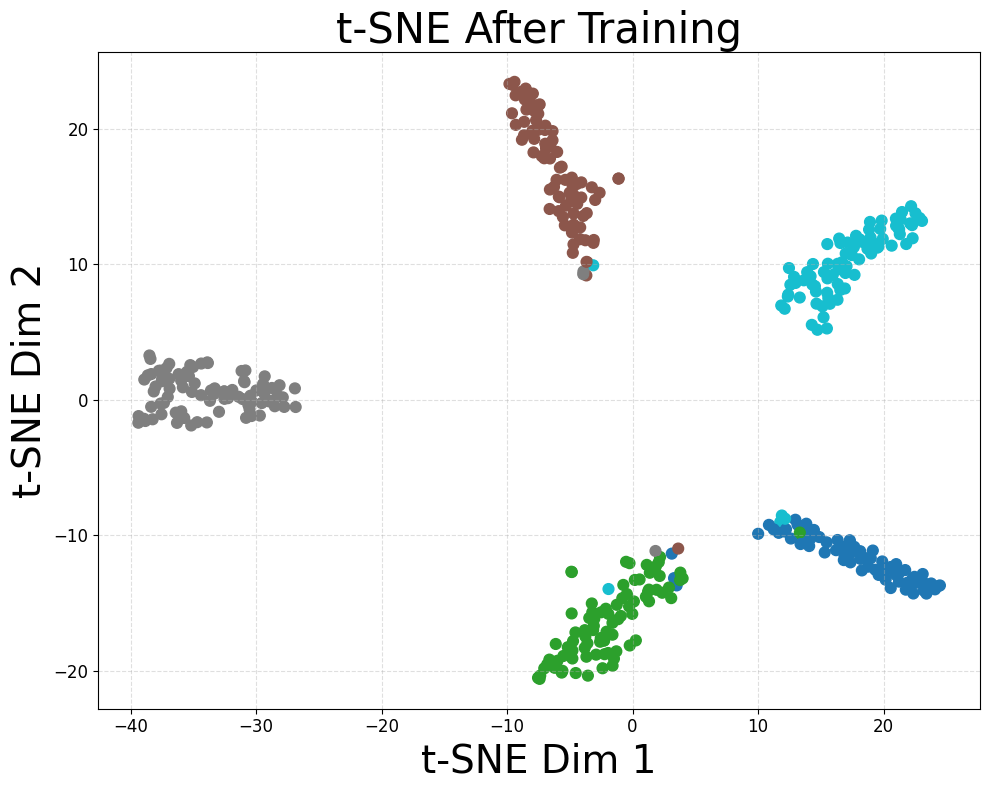}
        \label{fig:Pred_SLD-1_40}}
    \hfill
    \subfigure[50\% labeled]{
        \includegraphics[width=0.31\textwidth, height=3.5cm]{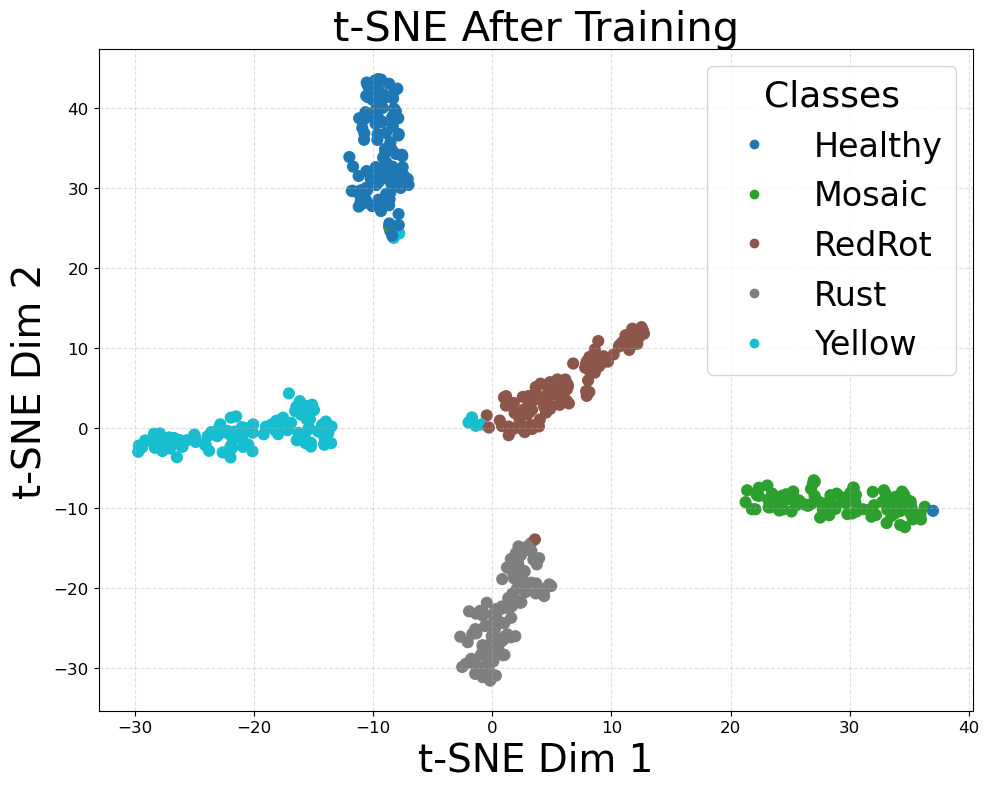}
        \label{fig:Pred_SLD-1_50}}
    \caption{t-SNE representation of SLD-1 data with different labeling ratios.}
    \label{fig:results_SLD1}
\end{figure}

\autoref{fig:results_SLD1} presents the t-SNE visualization of data representations before and after model training with different proportions of labeled leaf samples. 
The first plot \autoref{fig:roc_SLD-1} shows the data before training. 
When 10\% of the data is labeled \autoref{fig:Pred_SLD-1_10}, different clusters begin to form, indicating improved organization of the feature space and better classification performance with fewer labeled samples. 
With 20\% labeled data \autoref{fig:Pred_SLD-1_20}, the separation between classes becomes more apparent, demonstrating further enhancement in feature discrimination.
The clusters become more compact, well separated, and show more powerful feature learning and clear class boundaries as the labeling ratio increases to 30\% labeled data (\autoref{fig:Pred_SLD-1_30})–50\%  labeled data (\autoref{fig:Pred_SLD-1_50}).

\subsection{Limitation}
While the proposed framework shows strong performance in sugarcane leaf disease detection, some limitations remain. Its generalizability to other crops or varied environmental conditions may be limited. The use of unlabeled data can introduce noise, mainly in images with distortions or overlapping symptoms. Moreover, accurately detecting multiple disease levels and distinguishing between visually similar classes remains challenging, thereby affecting overall classification performance.

\section{Conclusion}
\label{Conclusion}
This study presents a U-SSDNet approach for identifying sugarcane leaf diseases that combines the Mean-Teacher framework with wavelet-based preprocessing, KL-divergence-driven pseudo-label correction, and Supervised Contrastive Learning. The presented approach provides benefits for leaf disease identification. It reduces the need for a large annotated dataset by leveraging an unlabeled dataset, making it suitable for real agricultural conditions. The wavelet-based preprocessing improves lesion visibility and reduces divergence within disease classes. KL-divergence helps correct pseudo-labels, reduces confirmation bias, and yields more accurate predictions. Supervised contrastive learning further improves feature distinction and supports clearer separation between disease classes. An experimental analysis of the sugarcane leaf disease dataset shows that the presented model achieves performance levels comparable to those of fully supervised approaches with fewer labeled samples and outperforms existing methods. The results indicate that the method can serve as a practical and reliable approach for early disease detection and precise monitoring in agriculture.

\bibliographystyle{elsarticle-harv}
\bibliography{Reference_C}

@article{Barbedo_2016, title={A review on the main challenges in automatic plant disease identification based on visible range images}, volume={144}, ISSN={1537-5110},  DOI={10.1016/j.biosystemseng.2016.01.017}, journal={Biosystems Engineering}, publisher={Elsevier BV}, author={Barbedo, Jayme Garcia Arnal}, year={2016}, month=apr, pages={52–60} }

@article{Strachan_2022, title={Latent potential of current plant diagnostics for detection of sugarcane diseases}, volume={4}, ISSN={2590-2628}, DOI={10.1016/j.crbiot.2022.10.002}, journal={Current Research in Biotechnology}, publisher={Elsevier BV}, author={Strachan, Simon and Bhuiyan, Shamsul A. and Thompson, Nicole and Nguyen, Nam-Trung and Ford, Rebecca and Shiddiky, Muhammad J.A.}, year={2022}, pages={475–492} }

@article{Dey_2022, title={Comparative performance of four CNN-based deep learning variants in detecting Hispa pest, two fungal diseases, and NPK deficiency symptoms of rice (Oryza sativa)}, volume={202}, ISSN={0168-1699}, DOI={10.1016/j.compag.2022.107340}, journal={Computers and Electronics in Agriculture}, publisher={Elsevier BV}, author={Dey, Biplob and Masum Ul Haque, Mohammed and Khatun, Rahela and Ahmed, Romel}, year={2022}, month=nov, pages={107340} }

@article{Lu_2020, title={A survey of public datasets for computer vision tasks in precision agriculture}, volume={178}, ISSN={0168-1699}, DOI={10.1016/j.compag.2020.105760}, journal={Computers and Electronics in Agriculture}, publisher={Elsevier BV}, author={Lu, Yuzhen and Young, Sierra}, year={2020}, month=nov, pages={105760} }

@article{Barbedo_2018, title={Impact of dataset size and variety on the effectiveness of deep learning and transfer learning for plant disease classification}, volume={153}, ISSN={0168-1699},  DOI={10.1016/j.compag.2018.08.013}, journal={Computers and Electronics in Agriculture}, publisher={Elsevier BV}, author={Barbedo, Jayme Garcia Arnal}, year={2018}, month=oct, pages={46–53} }

@article{Benchallal_2024, title={ConvNeXt based semi-supervised approach with consistency regularization for weeds classification}, volume={239}, ISSN={0957-4174}, DOI={10.1016/j.eswa.2023.122222}, journal={Expert Systems with Applications}, publisher={Elsevier BV}, author={Benchallal, Farouq and Hafiane, Adel and Ragot, Nicolas and Canals, Raphaël}, year={2024}, month=apr, pages={122222} }

@article{Yilma_2025, title={Attentive Self-supervised Contrastive Learning (ASCL) for plant disease classification}, volume={25}, ISSN={2590-1230}, DOI={10.1016/j.rineng.2025.103922}, journal={Results in Engineering}, publisher={Elsevier BV}, author={Yilma, Getinet and Dagne, Mesfin and Ahmed, Mohammed Kemal and Bellam, Ravindra Babu}, year={2025}, month=mar, pages={103922} }

@article{Hu_2025, title={A semi-supervised detection method for multi-scale tea geometrid by integrating intra- and inter-class information in natural scene images}, volume={237}, ISSN={0168-1699}, DOI={10.1016/j.compag.2025.110645}, journal={Computers and Electronics in Agriculture}, publisher={Elsevier BV}, author={Hu, Gensheng and Zhao, Yunlong and Bao, Wenxia and Zhang, Dongyan}, year={2025}, month=oct, pages={110645} }

@article{Ai_2025, title={Semi-supervised contrastive learning for flotation process monitoring with uncertainty-aware prototype optimization}, volume={145}, ISSN={0952-1976}, DOI={10.1016/j.engappai.2025.110222}, journal={Engineering Applications of Artificial Intelligence}, publisher={Elsevier BV}, author={Ai, Mingxi and Zhang, Jin and Li, Peng and Wu, Jiande and Tang, Zhaohui and Xie, Yongfang}, year={2025}, month=apr, pages={110222} }

@article{Sharma_2025, title={SoyaTrans: A novel transformer model for fine-grained visual classification of soybean leaf disease diagnosis}, volume={260}, ISSN={0957-4174}, DOI={10.1016/j.eswa.2024.125385}, journal={Expert Systems with Applications}, publisher={Elsevier BV}, author={Sharma, Vivek and Tripathi, Ashish Kumar and Mittal, Himanshu and Nkenyereye, Lewis}, year={2025}, month=jan, pages={125385}}

@article{Shah_2022, title={ResTS: Residual Deep interpretable architecture for plant disease detection}, volume={9}, ISSN={2214-3173}, DOI={10.1016/j.inpa.2021.06.001}, number={2}, journal={Information Processing in Agriculture}, publisher={Elsevier BV}, author={Shah, Dhruvil and Trivedi, Vishvesh and Sheth, Vinay and Shah, Aakash and Chauhan, Uttam}, year={2022}, month=jun, pages={212–223} }

@article{Ye_2024, title={Application of Tswin-F network based on multi-scale feature fusion in tomato leaf lesion recognition}, volume={156}, ISSN={0031-3203}, DOI={10.1016/j.patcog.2024.110775}, journal={Pattern Recognition}, publisher={Elsevier BV}, author={Ye, Yuanbo and Zhou, Houkui and Yu, Huimin and Hu, Haoji and Zhang, Guangqun and Hu, Junguo and He, Tao}, year={2024}, month=dec, pages={110775} }

@article{Ilsever_2024, title={Consistency regularization based semi-supervised plant disease recognition}, volume={9}, ISSN={2772-3755}, DOI={10.1016/j.atech.2024.100613}, journal={Smart Agricultural Technology}, publisher={Elsevier BV}, author={Ilsever, Murat and Baz, Ipek}, year={2024}, month=dec, pages={100613} }

@article{Zhang_2024, title={MAIANet: Signal modulation in cassava leaf disease classification}, volume={225}, ISSN={0168-1699}, DOI={10.1016/j.compag.2024.109351}, journal={Computers and Electronics in Agriculture}, publisher={Elsevier BV}, author={Zhang, Jiayu and Zhang, Baohua and Qi, Chao and Nyalala, Innocent and Mecha, Peter and Chen, Kunjie and Gao, Junfeng}, year={2024}, month=oct, pages={109351} }

@article{Thite_2024, title={Sugarcane leaf dataset: A dataset for disease detection and classification for machine learning applications}, volume={53}, ISSN={2352-3409}, DOI={10.1016/j.dib.2024.110268}, journal={Data in Brief}, publisher={Elsevier BV}, author={Thite, Sandip and Suryawanshi, Yogesh and Patil, Kailas and Chumchu, Prawit}, year={2024}, month=apr, pages={110268} }

@inproceedings{Lambor_2022, title={Sugarcane Leaf Disease Classification using Transfer Learning}, DOI={10.1109/iatmsi56455.2022.10119309}, booktitle={2022 IEEE Conference on Interdisciplinary Approaches in Technology and Management for Social Innovation (IATMSI)}, publisher={IEEE}, author={Lambor, Shilpa and Pungliya, Vithika and Bhonsle, Roshita and Purohit, Atharva and Raut, Ankur and Patel, Aayushi}, year={2022}, month=dec, pages={1–4} }

@inproceedings{He_2016, title={Deep Residual Learning for Image Recognition}, DOI={10.1109/cvpr.2016.90}, booktitle={2016 IEEE Conference on Computer Vision and Pattern Recognition (CVPR)}, publisher={IEEE}, author={He, Kaiming and Zhang, Xiangyu and Ren, Shaoqing and Sun, Jian}, year={2016}, month=jun, pages={770–778} }

@inproceedings{Liu_2015, title={Very deep convolutional neural network based image classification using small training sample size}, DOI={10.1109/acpr.2015.7486599}, booktitle={2015 3rd IAPR Asian Conference on Pattern Recognition (ACPR)}, publisher={IEEE}, author={Liu, Shuying and Deng, Weihong}, year={2015}, month=nov, pages={730–734} }

@article{Huang_2017, title={Densely connected convolutional networks}, author={Huang, Gao and Liu, Zhuang and Van Der Maaten, Laurens and Weinberger, Kilian Q.}, journal={Proceedings of the IEEE Conference on Computer Vision and Pattern Recognition (CVPR)}, year={2017}, publisher={IEEE}, pages={4700--4708}, }

@article{Dosovitskiy_2020, title={An image is worth 16x16 words: Transformers for image recognition at scale}, author={Dosovitskiy, Alexey and Beyer, Lucas and Kolesnikov, Alexander and Weissenborn, Dirk and Zhai, Xiaohua and Unterthiner, Thomas and Dehghani, Mostafa and Minderer, Matthias and Heigold, Georg and Gelly, Sylvain and Uszkoreit, Jakob and Houlsby, Neil}, journal={arXiv preprint arXiv:2010.11929}, publisher={arXiv}, year={2020}, month=oct, }

@inproceedings{Liu_2021, title={Swin Transformer: Hierarchical Vision Transformer using Shifted Windows}, DOI={10.1109/iccv48922.2021.00986}, booktitle={2021 IEEE/CVF International Conference on Computer Vision (ICCV)}, publisher={IEEE}, author={Liu, Ze and Lin, Yutong and Cao, Yue and Hu, Han and Wei, Yixuan and Zhang, Zheng and Lin, Stephen and Guo, Baining}, year={2021}, month=oct, pages={9992–10002}}

@article{Tarvainen_2017, title={Mean teachers are better role models: Weight-averaged consistency targets improve semi-supervised deep learning results}, author={Tarvainen, Antti and Valpola, Harri},
journal={Advances in Neural Information Processing Systems}, volume={30}, year={2017}, publisher={NeurIPS}, }

@InProceedings{Panigrahi_2020, author="Panigrahi, Kshyanaprava Panda and Das, Himansu and Sahoo, Abhaya Kumar and Moharana, Suresh Chandra", editor="Das, Himansu and Pattnaik, Prasant Kumar and Rautaray, Siddharth Swarup and Li, Kuan-Ching", title="Maize Leaf Disease Detection and Classification Using Machine Learning Algorithms", booktitle="Progress in Computing, Analytics and Networking", year="2020", publisher="Springer Singapore", address="Singapore",
pages="659--669", DOI={10.1007/978-981-15-2414-1_66}, isbn="978-981-15-2414-1"}

@article{Mia_2020, title={Mango leaf disease recognition using neural network and support vector machine}, volume={3}, ISSN={2520-8446}, DOI={10.1007/s42044-020-00057-z}, number={3}, journal={Iran Journal of Computer Science}, publisher={Springer Science and Business Media LLC}, author={Mia, Md. Rasel and Roy, Sujit and Das, Subrata Kumar and Rahman, Md. Atikur}, year={2020}, month=apr, pages={185–193} }

@inproceedings{Sari_2021, title={Classification of Chili Leaf Disease Using the Gray Level Co-occurrence Matrix (GLCM) and the Support Vector Machine (SVM) Methods}, DOI={10.1109/icic54025.2021.9632920}, booktitle={2021 Sixth International Conference on Informatics and Computing (ICIC)}, publisher={IEEE}, author={Sari, Yuslena and Baskara, Andreyan Rizky and Wahyuni, Rika}, year={2021}, month=nov, pages={1–4}}

@article{Dyrmann_2016, title={Plant species classification using deep convolutional neural network}, volume={151}, ISSN={1537-5110}, DOI={10.1016/j.biosystemseng.2016.08.024}, journal={Biosystems Engineering}, publisher={Elsevier BV}, author={Dyrmann, Mads and Karstoft, Henrik and Midtiby, Henrik Skov}, year={2016}, month=nov, pages={72–80} }

@article{Mohanty_2016, title={Using Deep Learning for Image-Based Plant Disease Detection}, volume={7}, ISSN={1664-462X}, DOI={10.3389/fpls.2016.01419}, journal={Frontiers in Plant Science}, publisher={Frontiers Media SA}, author={Mohanty, Sharada P. and Hughes, David P. and Salathé, Marcel}, year={2016}, month=sep }

@article{Tang_2020, title={Grape disease image classification based on lightweight convolution neural networks and channelwise attention}, volume={178}, ISSN={0168-1699}, DOI={10.1016/j.compag.2020.105735}, journal={Computers and Electronics in Agriculture}, publisher={Elsevier BV}, author={Tang, Zhe and Yang, Jialing and Li, Zhe and Qi, Fang}, year={2020}, month=nov, pages={105735} }

@article{Ali_2024, title={AppleLeafNet: a lightweight and efficient deep learning framework for diagnosing apple leaf diseases}, volume={15}, ISSN={1664-462X}, DOI={10.3389/fpls.2024.1502314}, journal={Frontiers in Plant Science}, publisher={Frontiers Media SA}, author={Ali, Muhammad Umair and Khalid, Majdi and Farrash, Majed and Lahza, Hassan Fareed M. and Zafar, Amad and Kim, Seong-Han}, year={2024}, month=nov }

@inproceedings{Kavala_2022, title={Detection Of Grape Leaf Disease Using Transfer Learning Methods: VGG16 \&amp; VGG19}, DOI={10.1109/iccmc53470.2022.9753773}, booktitle={2022 6th International Conference on Computing Methodologies and Communication (ICCMC)}, publisher={IEEE}, author={Kavala, Sri Adi Pavan Naidu and Pothuraju, Rajarajeswari}, year={2022}, month=mar,  pages={1205--1208} }

@article{Liu_2022, title={SK-MobileNet: A Lightweight Adaptive Network Based on Complex Deep Transfer Learning for Plant Disease Recognition}, volume={48}, ISSN={2191-4281}, DOI={10.1007/s13369-022-06987-z}, number={2}, journal={Arabian Journal for Science and Engineering}, publisher={Springer Science and Business Media LLC}, author={Liu, Guangsheng and Peng, Jialiang and El-Latif, Ahmed A. Abd}, year={2022}, month=jun, pages={1661–1675} }

@article{Singh_2022, title={Deep transfer modeling for classification of Maize Plant Leaf Disease}, volume={81}, ISSN={1573-7721}, DOI={10.1007/s11042-021-11763-6}, number={5}, journal={Multimedia Tools and Applications}, publisher={Springer Science and Business Media LLC}, author={Singh, Rajeev Kumar and Tiwari, Akhilesh and Gupta, Rajendra Kumar}, year={2022}, month=jan, pages={6051–6067} }

@inproceedings{Chen_2020, title={A Simple Framework for Contrastive Learning of Visual Representations}, author={Chen, Ting and Kornblith, Simon and Norouzi, Mohammad and Hinton, Geoffrey}, booktitle={International Conference on Machine Learning}, publisher={PMLR}, year={2020}, pages={1597--1607} }

@article{Xu_2023, title={A Comprehensive Survey of Image Augmentation Techniques for Deep Learning}, volume={137}, ISSN={0031-3203}, DOI={10.1016/j.patcog.2023.109347}, journal={Pattern Recognition}, publisher={Elsevier BV}, author={Xu, Mingle and Yoon, Sook and Fuentes, Alvaro and Park, Dong Sun}, year={2023}, month=may, pages={109347} }

@article{Naveed_2024, title={Survey: Image mixing and deleting for data augmentation}, volume={131}, ISSN={0952-1976}, DOI={10.1016/j.engappai.2023.107791}, journal={Engineering Applications of Artificial Intelligence}, publisher={Elsevier BV}, author={Naveed, Humza and Anwar, Saeed and Hayat, Munawar and Javed, Kashif and Mian, Ajmal}, year={2024}, month=may, pages={107791} }

@article{Lu_2023, title={Uncertainty-aware pseudo-label and consistency for semi-supervised medical image segmentation}, volume={79}, ISSN={1746-8094}, DOI={10.1016/j.bspc.2022.104203}, journal={Biomedical Signal Processing and Control}, publisher={Elsevier BV}, author={Lu, Liyun and Yin, Mengxiao and Fu, Liyao and Yang, Feng}, year={2023}, month=jan, pages={104203} }

@article{Huang_2025, title={GapMatch: Bridging Instance and Model Perturbations for Enhanced Semi-Supervised Medical Image Segmentation}, volume={39}, ISSN={2159-5399}, DOI={10.1609/aaai.v39i16.33919}, number={16}, journal={Proceedings of the AAAI Conference on Artificial Intelligence}, publisher={Association for the Advancement of Artificial Intelligence (AAAI)}, author={Huang, Wei and Zhang, Lei and Wang, Zizhou and Wang, Yan}, year={2025}, month=apr, pages={17458–17466} }

@article{Dangi_2025, title={A multi-temporal multi-spectral attention-augmented deep convolution neural network with contrastive learning for crop yield prediction}, volume={239}, ISSN={0168-1699}, DOI={10.1016/j.compag.2025.110895}, journal={Computers and Electronics in Agriculture}, publisher={Elsevier BV}, author={Dangi, Shalini and Mullapudi, Surya Karthikeya and Raghaw, Chandravardhan Singh and Dar, Shahid Shafi and Rehman, Mohammad Zia Ur and Kumar, Nagendra}, year={2025}, month=dec, pages={110895} }

@article{Liu_2024, title={ASCL: Accelerating semi‐supervised learning via contrastive learning}, volume={36}, ISSN={1532-0634}, DOI={10.1002/cpe.8293}, number={28}, journal={Concurrency and Computation: Practice and Experience}, publisher={Wiley}, author={Liu, Haixiong and Li, Zuoyong and Wu, Jiawei and Zeng, Kun and Hu, Rong and Zeng, Wei}, year={2024}, month=oct }

@article{Paramanandham_2024, title={Enhancing Disease Detection With Weight Initialization and Residual Connections Using LeafNet for Groundnut Leaf Diseases}, volume={12}, ISSN={2169-3536}, DOI={10.1109/access.2024.3422311}, journal={IEEE Access}, publisher={Institute of Electrical and Electronics Engineers (IEEE)}, author={Paramanandham, Nirmala and Sundhar, Shyam and Priya, P.}, year={2024}, pages={91511–91526} }

@inproceedings{Awais_2025, title={AgroGPT: Efficient Agricultural Vision-Language Model with Expert Tuning}, DOI={10.1109/wacv61041.2025.00555}, booktitle={2025 IEEE/CVF Winter Conference on Applications of Computer Vision (WACV)}, publisher={IEEE}, author={Awais, Muhammad and Salem Abdulla Alharthi, Ali Husain and Kumar, Amandeep and Cholakkal, Hisham and Anwer, Rao Muhammad}, year={2025}, month=feb, pages={5687–5696} }

@inproceedings{Singh_2024, title={RedRot and Mosaic Sugarcane Leaf Disease Classification using an EfficientNet Deep Learning Model}, DOI={10.1109/aic61668.2024.10731071}, booktitle={2024 IEEE 3rd World Conference on Applied Intelligence and Computing (AIC)}, publisher={IEEE}, author={Singh, Gurpreet and Guleria, Kalpna and Sharma, Shagun}, year={2024}, month=jul, pages={1058–1063} }

@inproceedings{Anika_2025, title={Classification and Identification of Saccharum Officinarum Diseases using Deep Transfer Learning Computer Vision}, DOI={10.1109/acit65614.2025.11185643}, booktitle={2025 15th International Conference on Advanced Computer Information Technologies (ACIT)}, publisher={IEEE}, author={Anika, Salma and Alon, Alvin Sarraga and Alejandro, Leonard L. and Vergara, Charlene I. and Paguigan, Jesus}, year={2025}, month=sep, pages={927–932} }

@inproceedings{Nomani_2024, series={ICCA 2024}, title={Sugarcane Leaf Disease Detection: A Comparative Analysis Using Deep Learning}, DOI={10.1145/3723178.3723197}, booktitle={Proceedings of the 3rd International Conference on Computing Advancements}, publisher={ACM}, author={Nomani, Khan Md Shibli and Nuruzzaman, Md and Nadia, Afsana and Billal, Md Masum}, year={2024}, month=oct, pages={139–144}, collection={ICCA 2024} }

@inproceedings{Sohn_2020_FixMatch, title = {FixMatch: Simplifying Semi-Supervised Learning with Consistency and Confidence}, author = {Sohn, Kihyuk and Berthelot, David and Li, Chun-Liang and Zhang, Zizhao and Carlini, Nicholas and Cubuk, Ekin D. and Kurakin, Alex and Zhang, Han and Raffel, Colin}, booktitle = {Advances in Neural Information Processing Systems}, volume = {33},
pages = {596-608}, year = {2020}, publisher = {Curran Associates, Inc.}}

@inproceedings{Zhang_2021_FlexMatch, title = {FlexMatch: Boosting Semi-Supervised Learning with Curriculum Pseudo Labeling}, author = {Zhang, Bowen and Wang, Yidong and Hou, Wenxin and Wu, Hao and Wang, Jindong and Okumura, Manabu and Shinozaki, Takahiro}, booktitle = {Advances in Neural Information Processing Systems}, volume = {34}, pages = {18408--18419}, year = {2021},
publisher = {Curran Associates, Inc.}}

@inproceedings{Chen_2023_SoftMatch, title = {SoftMatch: Addressing the Quantity-Quality Trade-off in Semi-supervised Learning}, author = {Chen, Hao and Tao, Ran and Fan, Yue and Wang, Yidong and Wang, Jindong and Schiele, Bernt and Xie, Xing and Raj, Bhiksha and Savvides, Marios}, booktitle = {Proceedings of the International Conference on Learning Representations}, year = {2023}}

@article{Rehman_2023, title={KisanQRS: A deep learning-based automated query-response system for agricultural decision-making}, volume={213}, ISSN={0168-1699}, DOI={10.1016/j.compag.2023.108180}, journal={Computers and Electronics in Agriculture}, publisher={Elsevier BV}, author={Rehman, Mohammad Zia Ur and Raghuvanshi, Devraj and Kumar, Nagendra}, year={2023}, month=oct, pages={108180} }

@article{Raghaw_2026, title={H-SemiS: Hierarchical fusion of semi and self-supervised learning for knee osteoarthritis severity grading}, volume={322}, ISSN={0957-4174},  DOI={10.1016/j.eswa.2026.132279}, journal={Expert Systems with Applications}, publisher={Elsevier BV}, author={Raghaw, Chandravardhan Singh and Parwal, Anushka and Dar, Shahid Shafi and Darade, Prajakta and Kumar, Nagendra}, year={2026}, month=Aug, pages={132279} }

@article{Raghaw_2024, title={An Explainable Contrastive-based Dilated Convolutional Network with Transformer for Pediatric Pneumonia Detection}, volume={167}, ISSN={1568-4946}, DOI={10.1016/j.asoc.2024.112258}, journal={Applied Soft Computing}, publisher={Elsevier BV}, author={Raghaw, Chandravardhan Singh and Bhore, Parth Shirish and Rehman, Mohammad Zia Ur and Kumar, Nagendra}, year={2024}, month=Dec, pages={112258} }
\end{document}